\documentclass[sigconf]{acmart}
\acmConference[KDD '25]{Make sure to enter the correct conference title from your rights confirmation email}{August 3--7, 2025}{Toronto, Canada}
\usepackage{amsmath}
\usepackage{dsfont}
\usepackage{algorithm}
\usepackage{algpseudocode}
\usepackage{amsmath}

\AtBeginDocument{%
  }

\copyrightyear{2025}
\acmYear{2025}
\setcopyright{acmlicensed}\acmConference[KDD '25]{Proceedings of the 31st ACM SIGKDD Conference on Knowledge Discovery and Data Mining V.1}{August 3--7, 2025}{Toronto, ON, Canada}
\acmBooktitle{Proceedings of the 31st ACM SIGKDD Conference on Knowledge Discovery and Data Mining V.1 (KDD '25), August 3--7, 2025, Toronto, ON, Canada}
\acmDOI{10.1145/3690624.3709309}
\acmISBN{979-8-4007-1245-6/25/08}

\begin{document}

\title{BTFL: A Bayesian-based Test-Time Generalization
Method for Internal and External Data Distributions
in Federated learning}

\author{Yu Zhou}
\email{zhouyucs@bupt.edu.cn}
\affiliation{%
    \institution{Beijing University of Posts and Telecommunications}
    \city{Beijing}
    \country{China}
}

\author{Bingyan Liu}
\authornote{Corresponding author.}
\email{bingyanliu@bupt.edu.cn}
\affiliation{%
     \institution{Beijing University of Posts and Telecommunications}
    \city{Beijing}
    \country{China}
}

\begin{abstract}
  Federated Learning (FL) enables multiple clients to collaboratively develop a global model while maintaining data privacy. However, online FL deployment faces challenges due to distribution shifts and evolving test samples. Personalized Federated Learning (PFL) tailors the global model to individual client distributions, but struggles with Out-Of-Distribution (OOD) samples during testing, leading to performance degradation. In real-world scenarios, balancing personalization and generalization during online testing is crucial and existing methods primarily focus on training-phase generalization. To address the test-time trade-off, we introduce a new scenario: Test-time Generalization for Internal and External Distributions in Federated Learning (TGFL), which evaluates adaptability under Internal Distribution (IND) and External Distribution (EXD). We propose \textbf{BTFL}, a Bayesian-based test-time generalization method for TGFL, which balances generalization and personalization at the sample level during testing. BTFL employs a two-head architecture to store local and global knowledge, interpolating predictions via a dual-Bayesian framework that considers both historical test data and current sample characteristics with theoretical guarantee and faster speed. Our experiments demonstrate that BTFL achieves improved performance across various datasets and models with less time cost. The source codes are made publicly available at \href{https://github.com/ZhouYuCS/BTFL}{https://github.com/ZhouYuCS/BTFL
}.
\end{abstract}


 
\maketitle

\section{Introduction}
\label{sec:intro}
Federated Learning (FL) is a prevalent approach where multiple clients jointly develop a global model while maintaining data privacy~\cite{mcmahan2017communication,wang2023dafkd,wang2024feddse,liu2024recent}. However, online FL deployment can be challenging due to potential distribution shifts and evolving test samples. Personalized Federated Learning (PFL) can be used to tailor the global model to the specific distribution of individual clients~\cite{wang2019federated,liu2021pfa,liu2021distfl,liu2023beyond}. Despite effectiveness, its application is limited in real-world scenarios where Out-Of-Distribution (OOD) samples~\cite{recht2018cifar,hendrycks2018benchmarking,wang2024traceable} come during the test time. For example, in the context of species image classification, if one client aims to participate in FL in winter, it may capture rare snake records for local training due to hibernation. Then, once the client deploys the model with personalization and spring arrives, not only does the environmental background change, but snakes also emerge and are detected by the client's model. These new scenes/species, being OOD, can potentially confuse the excessively personalized local model and make it hard to generalize. Therefore, it is crucial to find a balance between personalization and the generalization capabilities of local models during online testing in the context of FL.


There are a number of work targeting FL generalization. For example, Caldarola \textit{et al.}~\cite{caldarola2022improving} proposed an FL generalization method by seeking the flat minimum value of the loss function. FedSR~\cite{nguyen2022fedsr} improves the model's generalization ability for unknown target data based on L2 regularization and conditional mutual information regularization to learn simple and robust representations. However, existing work pays more attention to generalization in FL at the training phase. For the test-time adaptation, a closely related work is FedTHE, where the authors tried to get robust test-time personalization by ensembling personal head and global head~\cite{jiang2022test}. After exploration, we discover that FedTHE is empirical and can easily lead to prejudice when facing challenging generalization scenarios (details in Section~\ref{sec:motivation}).

To the best of our knowledge, there is no existing work that can achieve test-time generalization for various distribution conditions in FL with theoretical guarantee. To fill the gap, we first introduce a new scenario, named \textit{Test-time Generalization for Internal and External Distributions in Federated learning (TGFL)}, where we assume samples at the test time conform to either an Internal Distribution(IND), which means that the sample is independently and identically distributed with the local training set, or an External Distribution(EXD), which refers to samples from external clients with new labels to the local personalized models. Compared to the standard FL, TGFL can be used to achieve a more comprehensive evaluation for the adaptability when distribution conditions change.
Following this scenario, we propose \textbf{BTFL}, a \textbf{B}ayesian-based \textbf{T}est-time generalization method for TG\textbf{FL}, which achieves desirable performance under both IND and EXD scenarios. The \emph{key idea} of BTFL is to leverage Bayesian methods to explore the optimal trade-off between knowledge with strong generalization to EXD and knowledge inclined to personalization on IND at the sample-level during the test time.






Specifically, BTFL first extracts and memorizes knowledge suited for IND/EXD, and continuously updates its knowledge selection strategy during the test time. Here we store the local and global knowledge through a two-head architecture, where one head is a generalized federated global classifier suited for EXD data, and the other is a personalized classifier solely trained on IND data. Both heads are obtained and kept on the client side when training produce is finished. With this architecture, BTFL extracts knowledge from a probability view, resulting in interpolation of predictions made by two heterogeneous heads, which is also used in earlier methods such as FedTHE~\cite{jiang2022test}. However, different from FedTHE that relies on unstable metrics (details in Section~\ref{sec:motivation}), in our work, the interpolation is achieved by a dual-Bayesian framework, which fully considers both the time series information about the recent test data flow, and the specific characteristic of the current sample itself with formal theoretical guarantee and faster speed. We elaborate the distinctive features of BTFL as follows:



(1) \textit{Theoretical Support.} 
We model the adaptation task under IND/EXD shifts as a posterior estimation problem. Based on this perspective, we treat knowledge from two heads as conditional probabilities under these shifts and propose an analytical solution grounded in Bayesian statistical theory, which can be  applied to both data flow information extraction and the analysis of current test samples, ensuring explainable decision-making processes.

(2) \textit{Optimization Free.} 
Compared to existing TTA methods, including those for FL scenarios, BTFL eliminates the need for additional optimization of model or other parameters (e.g., forward/backward pass). This is due to its knowledge extraction design with an analytical solution under the proposed probability view, which is equivalent to numerical integration on the aspect of computation. Consequently, many existing software solvers, such as \texttt{scipy.integrate} and \texttt{numpy.trapz}, can be effectively applied to accelerate BTFL, making it faster than other optimization-based methods.

Since existing benchmarks cannot fully evaluate the balance between IND and EXD, we manually construct one for more comprehensive evaluation (details in Section \ref{sec:ES}). Based on the benchmark, BTFL achieves both improved and interpretative effects across various datasets (CIFAR10, OfficeHome, and ImageNet) and models (CNN,
ResNet and Transformer). In addition, as a test-time method, BTFL achieves good performance without costly training processes compared to existing methods, which is especially beneficial for resource-constrained client devices to participate in FL. On average, BTFL surpasses other baselines by 1.88\%-3.15\% in accuracy with roughly 4X-12X time saving.

This paper makes the following contributions:

\begin{itemize}
    \item \textbf{A New Scenario and Benchmark.}
    We consider TGFL, a new learning setting in FL, targeting the challenge of generalizing to the external distribution while maintaining personalization on the internal distribution during the test time. We also propose an improved benchmark BTGFL for fair evaluation in the realistic FL scenario.
    
    
    \item \textbf{A Bayesian-based Method.}
    We propose BTFL, which leverages the real-world data statistical laws with Bayesian formula to achieve the optimal knowledge extraction of the global and local head.
     With the novel probability perspective, BTFL proposes an analytical method to avoid costly optimization-based optimization. As far as we know, BTFL is the first method that introduces probability perspective and Bayesian methods for test-time adaptation in FL. 

    \item  \textbf{Theoretical Guarantee.}
    BTFL is grounded in interpolation prediction by two heads, which balances the trade-off between personalization and generalization. It leverages the Beta-Bernoulli process for historical information analysis, and approximate Bayesian update for sample-level information process, to provide reliable parameter estimation for the derived analytical solution under a novel probability view. The proof details can be found in Appendix \ref{sec:bbp}

    \item  \textbf{Comprehensive Evaluation.}
     Experiments on multiple datasets, models and FL settings have shown that our method outperforms other baselines under BTGFL, with decreased time overhead.
    

\end{itemize}

\section{Related Work}
\subsection{Generalized Federated Learning}
In the field of FL, improving model generalization during training has attracted consistent interest~\cite{shi2021fed,qu2022generalized,wang2024fedcda,wang2024fednlr}. For example, FedSR~\cite{nguyen2022fedsr} used domain generalization techniques to enhance model performance for unseen data.  Caldarola~\cite{caldarola2022improving} proposed the use of Sharpness-Aware Minimization at client sites and Stochastic Weight Averaging at the server, demonstrating significant improvements in model generalization across heterogeneous and homogeneous scenarios. FLRA~\cite{reisizadeh2020robust} developed the FedRobust algorithm, a gradient descent ascent method to guard against affine distribution shifts. Although these methods can enhance the ability of model generalization in FL, all of them focus on the training phase, failing to satisfy an online adaptation during the test time. 

\subsection{Test-time Adaptation}
Test-Time Adaptation (TTA) methods have gained significant attention for enhancing model accuracy on out-of-distribution data without additional training data. These methods can be categorized based on whether they address \textit{feature shifts}, where the feature distribution changes (e.g., Tent~\cite{wang2020tent} and MEMO~\cite{zhang2021memo}), or label shift, where the label distribution changes (e.g., EM~\cite{alexandari2020maximum,saerens2002adjusting} and BBSE~\cite{azizzadenesheli2019regularized}). However, these methods may require access to the entire test dataset, which could be restrictive in online deployment scenarios. Recently, research on TTA methods designed for FL emerges. ATP~\cite{bao2024adaptive} adopted dynamic rate for test-time adaptation on novel clients; FedICON~\cite{tan2023taming} tried to guide adaptation through invariant features; FedTHE~\cite{jiang2022test} introduced a unique two-head design that learns global and local objectives and optimized the interpolation weight of the heads in the test time. In our work, we target TTA in FL from a probability perspective, which is significantly different from others and achieves improved results with less overhead and more theory support.

\subsection{Bayesian Methods in FL}
Bayesian methods have emerged as a potent strategy to enhance performance and robustness in various areas~\cite{zeng2024collapsed,seligmann2024beyond}. $\beta$-Predictive Bayes~\cite{fischer2024federated} introduced a novel perspective by interpolating between predictive posterior distributions with a tunable parameter $\beta$ and applied knowledge distillation to generate an improved single model. By leveraging unlabeled data for knowledge distillation, the nonparametric Bayesian~\cite{yurochkin2019bayesian}  allowed for the synthesis of a global network from local neural network weights with minimal communication rounds.  
In the field of FL, FedBE~\cite{chen2020fedbe} explored the application of Bayesian Deep Learning (BDL)~\cite{abdullah2024leveraging,wilson2020bayesian} models within FL, with a focus on the impact of different aggregation strategies on model performance. However, none of the works pay attention to apply Bayesian methods to the TTA problem under FL.

\section{Motivation}
\label{sec:motivation}

\begin{figure} 
  \centering
  \includegraphics[width=0.45\textwidth]{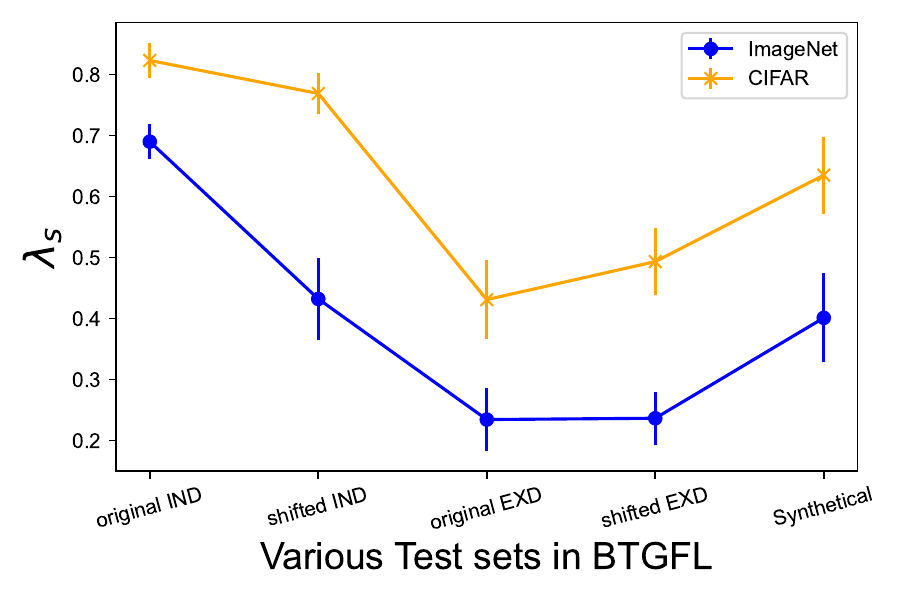} 
  \caption{Averaged $\lambda_{s}$ on different datasets with various distributions of our benchmark BTGFL.} 
  \label{fig:lambda} 
\end{figure}

\textbf{Limitation of current practice.}
As the first method that clarifies the dilemma between personalization and generalization during the test time, FedTHE~\cite{jiang2022test} provides a plausible solution to balance the degree. The core formula of FedTHE is 

\begin{equation}
\label{eq:slw}  
e^* = \arg\min_e \left( L_{SLW} := \lambda_s L_{EM} + (1 - \lambda_s)L_{FA} \right), 
\end{equation}
\[
\text{where} \quad \lambda_s = \cos(p(\hat{y}_g), p(\hat{y}_l)) \in [0, 1]  
\]
where $e^*$ is the interpolation coefficient between the personal head and global head. $L_{EM}$ indicates the Entropy Minimization Loss for the interpolated prediction.  $L_{FA}$ refers to the Feature Alignment Loss. $\lambda_s$ is the cosine similarity between the predicted logits made by the personal head $p(\hat{y}_l)$ and global head $p(\hat{y}_g)$.

A robust result comes from the balance between two loss items $\lambda_s L_{EM}$ and $(1 - \lambda_s)L_{FA}$. However, the balance is fragile when label space changes as verified by Figure~\ref{fig:lambda}. Although we expect that $\lambda_s$ will dynamically adjust the weight of two items, we discover that it always remains a lower value for the high-dimensional situation. What's more, the high-dimensional features naturally lead to higher value for $L_{FA}$, which further enhances the effects of $L_{FA}$ and decreases the relative influence of $L_{EM}$. As a result, the optimization objective in \eqref{eq:slw} varies as label dimension changes, which may mislead the optimization outside reasonable expectation.

Another pitfall of FedTHE is the use of Exponential-Moving-Average (EMA) to represent the recent test features. EMA typically contains excessively redundant information, and to eliminate this redundancy, it is necessary to average over a sufficiently large number of samples, utilizing the central limit theorem to filter out the noise. However, when the batch size decreases, the conditions for the central limit theorem are no longer satisfied, leading to the exposure of the noise and consequently a decline in performance (as shown in Table ~\ref{tab:EMA}).

\begin{table}
    \centering
    \captionof{table}{Accuracy of FedTHE with a $0.9$ EMA ratio and BTFL on batch-size of 1 and 32 under different distributions on CIFAR-10.}
    \label{tab:EMA}
    \begin{tabular}{l cccc } 
        \toprule
        & \parbox{1cm}{\centering Original \\ IND} & \parbox{0.9cm}{\centering Shifted \\ IND} & \parbox{0.9cm}{\centering Original \\ EXD} & \parbox{0.9cm}{\centering Shifted \\ EXD} \\
        \midrule
         FedTHE,bs=32 & 87.16\% & 74.95\% & 59.93\% & 46.32\%   \\
          FedTHE,bs=1 & 86.94\% & 74.14\% & 44.82\% & 40.29\%  \\ \midrule
          BTFL(Ours),bs=32 & 88.12\% & 79.74\% & 60.23\% & 48.31\%\\
         BTFL(Ours),bs=1 & 89.78\% & 80.00\% & 60.43\% & 48.97\%  \\
        \bottomrule
    \end{tabular}   
    \vspace{10pt}
\end{table}

\textbf{Intuition behind employing Bayesian methods.} Many work regard the adaptation task as the interpolation of original and fine-tuned classifier head, such as WiSE-FT~\cite{wortsman2022robust} and FedTHE~\cite{jiang2022test}. However, researches on this topic lack theory support and rely on unstable metrics as stated in the limitations. In our work, we attempt to fill this gap by rethinking interpolation in a probability perspective. 
Following ~\cite{wortsman2022robust}, we assume $\mathbf{V}$ are the parameters of the feature extractor, and 
$\mathbf{z}=f\left(x, \mathbf{V}\right)$ is the extracted public feature for heads. The head $C(\mathbf{z}, \theta)= \mathbf{z^{\top}} \theta$, where $\theta \in \mathbb{R}^{d \times k}$ is the classification matrix. For a mixing coefficient $m \in[0,1]$, we consider the output-space interpolation between the personalized head with parameters $\theta_{0}$ and the global head with parameters $\theta_{1}$. When personal head is obtained by fine-tuning the linear head, the output-space interpolation $osi$ is equivalent to the logits interpolation as follows.
\begin{equation}
\label{eq:ose}
osi(x;m,\mathbf{z})=(1-m)\cdot C(\mathbf{z};\theta_{0})+m \cdot C(\mathbf{z};\theta_{1})
\end{equation}

What inspires us is the form of Equation~\ref{eq:ose}, which is close to the probability equation if we consider $1-m$ and $m$ as prior while regarding $C(\mathbf{z},\theta_{0})$ and $C(\mathbf{z},\theta_{1})$ as the conditional probability. There is only one step left to make it a real probability equation: replace the logits with the final probability $\mathbf{y}$ by adding the $softmax$ operation. Similar to \eqref{eq:ose}, we can give the following prediction-space interpolation $psi$ as
\begin{equation}
\label{eq:psi}
psi(\mathbf{y} \mid m,\mathbf{z})=(1-m)\cdot Y(\mathbf{z}; \theta_{0})+m \cdot Y(\mathbf{z}; \theta_{1})
\end{equation}

A natural interpretation of \eqref{eq:psi} is that we have a prior that follows a binomial distribution $\mathcal{B}(n,m)$, with $m$ as a  parameter, standing for $P(x \sim \mathcal{EXD})$, the probability of $x$ conforming to the \textbf{External distribution} $\mathcal{EXD}$; $1-m$ represents the probability of the opposite event: $x$ conforms to the \textbf{Internal distribution} $\mathcal{IND}$, written as $P(x \sim \mathcal{IND})$; another parameter $n$ denotes the count of instances, referencing the preceding $n$ test samples;  $Y(\mathbf{z}; \theta_{0})$ and $Y(\mathbf{z}; \theta_{1})$ are the corresponding conditional probabilities $P(\mathbf{y}|x\sim \mathcal{IND})$ and $P(\mathbf{y}|x \sim \mathcal{EXD})$. With this perspective, we turn the interpolation problem into a parameter estimation problem for a binomial distribution $\mathcal{B}(n,m)$.

However, for a single sample $x$, we can only directly derive $\mathbf{z}$ as condition instead of the precise parameter $m$, which  drives our goal as \textit{discovering $psi(\mathbf{y} \mid \mathbf{z})$}. Fortunately, probability formula tells us that $psi(\mathbf{y} \mid \mathbf{z})$ is equivalent to 
\begin{equation}
\label{eq:prob}
\int  psi(\mathbf{y}  \mid \hat{m},\mathbf{z}) \cdot pdf(\hat{m}\mid \mathbf{z})\, d\hat{m}
\end{equation}
for any parameter $\hat{m}$, where $pdf(\hat{m}\mid \mathbf{z})$ is the conditional probability distribution function.

Since \eqref{eq:psi} takes advantage of knowledge from two heads to derive $psi(\mathbf{y} \mid m,\mathbf{z})$, there are two challenges for acquiring $pdf(\hat{m}\mid \mathbf{z})$ remained: On the one hand, we interpret $m$ as a parameter of a binomial distribution $\mathcal{B}(n,m)$, which should be estimated by time series samples without considering current $\mathbf{z}$, thus representing for historical information which is constantly updated as data flow continues; on the other hand, given prior probability distribution of $m$, we need further estimation of the posterior $pdf(\hat{m}\mid \mathbf{z})$ to work out $psi(\mathbf{y} \mid \mathbf{z})$ as implied in \eqref{eq:prob}. Both hands lead to application of Bayesian methods.



Besides theretical support, for the limitations of current practice, the Bayesian method can decouple the redundant information in the EMA-kind feature representation since we simplify historical information to the estimation of the parameter of a binomial distribution to extract past knowledge without EMA.
Also, the Bayesian method does not take cosine similarity into consideration to avoid dimensional difference shown in Figure ~\ref{tab:EMA}.

\begin{figure*}
  \centering
  \includegraphics[width=\textwidth]{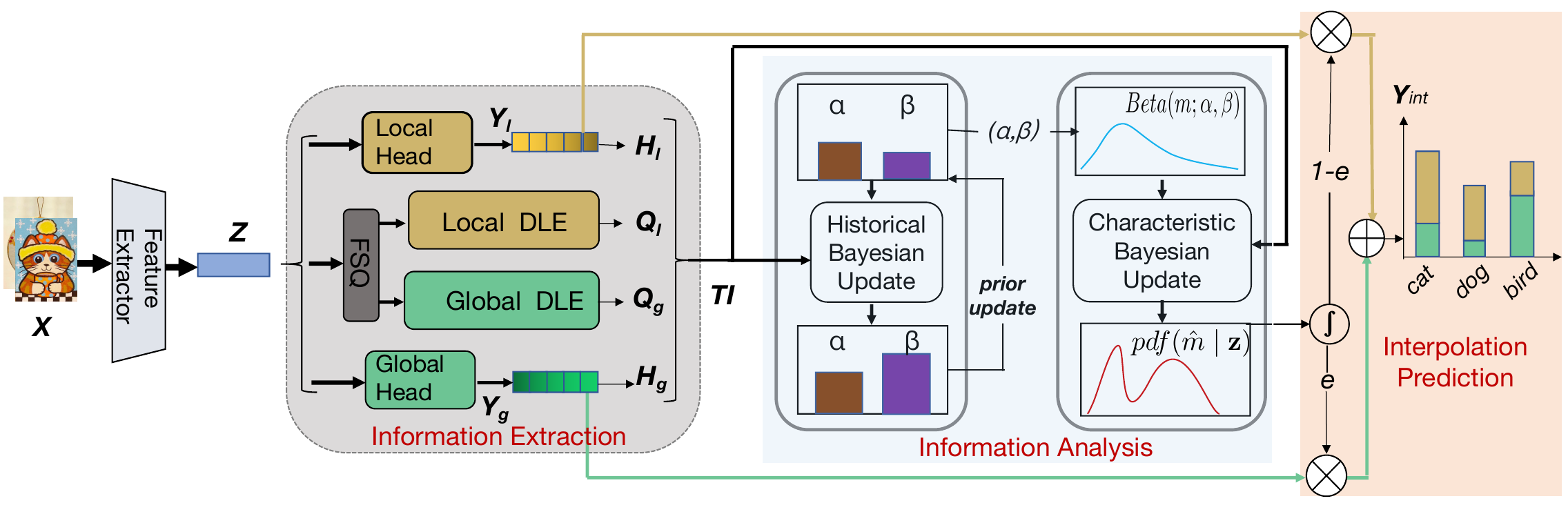} 
  \caption{ Overview of the proposed BTFL. It mainly includes three modules: Information Extraction, Information
Analysis and Interpolation Prediction.} 
  \label{fig:overview} 
\end{figure*}

\section{Framework of BTFL}
\subsection{Preliminary}
\textbf{Locally-kept Personal Head.}
As mentioned in the last section, we operate the head interpolation during the test time. To obtain the fine-tuned head to interpolate with the original global head $w_{g}$, we keep the feature extractor frozen and retrain a locally-kept personal head $w_{p}$ with local train set. Besides, we keep the averaged entropy of the two heads over local train set termed as $\overline{H_l}$ and $\overline{H_g}$ for future adaption.

\textbf{Discretised-Likelihood Estimators (DLE).}
Once training ends, each client is also required to derive the feature characteristic of their local dataset. Unlike FedTHE that averages features, we adopt a likelihood estimation method through discretization. Specifically, we compress each of the $d$ dimensions of feature $\mathbf{z}$ into a binary discrete space of $\{0, 1\}$ by the finite scalar quantization~\cite{mentzer2023finite}:
\begin{equation}
\label{eq:FSQ}
\hat{\mathbf{z}}=\mathrm{fsq}(\mathbf{z}):=\operatorname{Round}[ tanh(\mathbf{z})] 
\end{equation}
where $\mathrm{fsq}:\mathbb{R}^d \rightarrow {\{0,1\}}^d$ maps a continuous variable to a discretised space and makes it feasible to estimate the likelihood of $\mathbf{z}$ with that of $\hat{\mathbf{z}}$. The local discretised-Likelihood Estimator (DLE) preserves the frequency of the value being $0$ as $p_{i}$ for each dimension $i$ of the data features, formatting a $d$-dimensional parameter. When the model is well trained, each dimension of the feature tends to be independent. Therefore, for $\hat{\mathbf{z}} \in \{0, 1\}^{d}$ obtained by inputting feature $\mathbf{z}$ of sample $x$ into \eqref{eq:FSQ}, we can approximate $q_{dis}(\hat{\mathbf{z}}|x \sim \mathcal{IND})$, the discretised likelihood of $\hat{\mathbf{z}}$ under condition $x\sim \mathcal{IND}$ by: 
\begin{equation}
q_{dis}(\hat{\mathbf{z}}|x \sim \mathcal{IND}) \approx  \prod_{i=1}^{d} p_i^{\mathbb{I}(\hat{z}_i=0)} \cdot (1-p_i)^{\mathbb{I}(\hat{z}_i=1)}
\label{eq:DLE}
\end{equation}
Additionally, we need estimation of its counterpart $q_{dis}(\hat{\mathbf{z}}|x \sim \mathcal{EXD})$. Hence, each client uploads its local DLE to the server. The server then averages the  local DLEs to obtain global DLE, which is subsequently transmitted back to each client for estimation of $q_{dis}(\hat{\mathbf{z}}|x \sim \mathcal{EXD})$ like \eqref{eq:DLE}.

\subsection{Overview}
With above preliminaries, we give the overview of BTFL. As depicted in Figure~\ref{fig:overview}, to achieve our goal of obtaining  $Y_{int}:=psi(\mathbf{y} \mid \mathbf{z})$, BTFL mainly includes three modules to process the feature $\mathbf{z}$ of sample $x$ at test time:
\begin{itemize}
    \item \textit{Information Extraction.}
    There are four working units in this module: For each sample $x$, Local Head and Global Head use the feature $\mathbf{z}$ as input, separately outputs $P(\mathbf{y} \mid x \sim \mathcal{IND})$ notated as $Y_l$ and $P(\mathbf{y} \mid x \sim \mathcal{EXD})$ notated as $Y_g$; Local and Global discretised-likelihood Estimators (DLE) use the quantized feature $\hat{\mathbf{z}}$ obtained by \eqref{eq:FSQ} as input, separately outputs $Q_l$, a notation for $q_{dis}(\hat{\mathbf{z}}|x \sim \mathcal{IND})$ and $Q_g$, a notation for $q_{dis}(\hat{\mathbf{z}}|x \sim \mathcal{EXD})$. Besides, we keep each entropy of $Y_l$ and $Y_g$ termed as $H_l$ and $H_g$: Together with $Q_l$ and $Q_g$, they form the Test-Information ($TI$) for Information Analysis module.

    \item \textit{Information Analysis.}
    As mentioned in Section ~\ref{sec:motivation}, we build a Bayesian framework to solve the parameter estimation problem implied in \eqref{eq:psi}. Under Bayesian perspective, instead of direct estimation of $m$, we turn to the distribution of $m$, which is achieved by our Historical Bayesian Update (HBU) and characteristic Bayesian Update (CBU). HBU first analyzes occurrence of events through $TI$, and updates the prior of $m$ in a Beta-Bernoulli process~\cite{staton2018beta}; while by mining sample-level characteristic from $TI$, CBU further updates the prior of $m$ provided by HBU to derive $pdf(\hat{m} \mid \mathbf{z})$, the probability density function of $\hat{m}$ which is a random variable representing $m$ updated under characteristic-level event $\mathbf{z}$.

    \item \textit{Interpolation Prediction.}
    This module takes $Y_{l}$, $Y_{g}$ along with $pdf(\hat{m}\mid\mathbf{z})$ as input and outputs the adapted prediction $Y_{int}$ by interpolation of $Y_{l}$ and $Y_{g}$. This is done by our Dual Posterior Injection (DPI) with little calculation cost.
\end{itemize}

In the following parts, we will describe our key designs HBU, CBU and DPI in detail.

\subsection{Historical Bayesian Update (HBU)}

\textbf{Bayesian formula.}
HBU looks for the distribution of $m$, the parameter of $\mathcal{B}(n,m)$ illustrated in Section \ref{sec:motivation}, according to information hidden behind recent test flow.  Although we know nothing about the mechanism behind test distribution (i.e., source of the data stream), we can obtain $\mathrm{p}(m \mid \mathrm{\alpha,\beta})$ based on the Bayesian formula as follows:
\begin{equation}
\label{eq:Bayesian}
\mathrm{p}(m \mid \mathrm{\alpha,\beta})=\frac{\mathrm{p}(m) * \mathrm{q}(\mathrm{\alpha,\beta} \mid m)}{\mathrm{P}(\mathrm{\alpha,\beta})}
\end{equation}
where $\alpha$ is the number of samples that conform to $\mathcal{IND}$, $\beta$ is the number of samples that conform to $\mathcal{EXD}$ for recent flow; $\mathrm{p}(m)$ is the prior and $\mathrm{q}(\mathrm{\alpha,\beta} \mid m)$ is the likelihood; $\mathrm{P}(\alpha,\beta)$ is the evidence, and $\mathrm{p}(\mathrm{m} \mid \mathrm{\alpha,\beta})$ is the result of the prior update, which is the posterior. We initialize prior $\mathrm{p}(m)$ as $pdf$ of uniform distribution $\mathcal{U}(0,1)$ following the principle of insufficient reasons~\cite{dubs1942principle}. Once a new event of $x \sim \mathcal{IND}$ or $x \sim \mathcal{EXD}$ is detected, the prior is updated by \eqref{eq:Bayesian}. The update formula is analytically derived as follows:
\begin{equation}
\label{eq:beta}
\mathrm{p}(m \mid \mathrm{\alpha,\beta})=Beta(m;\alpha,\beta) 
\end{equation}
The result of formula \eqref{eq:beta} is deduced from the conjugate relationship between binomial distribution and beta distribution (Beta-Bernoulli process). Detailed theoretical proof can be found in Appendix \ref{sec:bbp}.


\textbf{Events detection.}
The technical difficulty of HBU lies in how to detect the new events for the Beta-Bernoulli process. Since $Q_l$ and $Q_g$ are inputs of HBU, detection could be done by comparing the ratio between $Q_l$ and $Q_g$ ,which is defined as:
\begin{equation}
\label{eq:tau}
\tau:=\frac{Q_l}{Q_g}
\end{equation}
However, $Q_g$ and $Q_l$ are just approximations of $q(\mathbf{z}|x \sim \mathcal{IND})$ and $q(\mathbf{z}|x \sim \mathcal{EXD})$, we cannot ensure the reliance of simply applying MLE, which may lead to failure in experiments. Instead, we mandate an inspection to determine whether samples should be considered for events detection from another perspective: as the lower bound of the empirical risk (see Appendix \ref{sec:lb}), entropy plays a significant role in reflecting the confidence level of a certain sample for the model~\cite{lee2024entropy,iwasawa2021test}. Thus we assume that if sample $ x\sim \mathcal{IND}$, $h_l$ will be at a lower level while $h_g$ will be relatively high, which introduces the second insurance to confirm the assumption of $ x\sim \mathcal{IND}$. Similarly, if Maximum Likelihood Estimation (MLE) implies $ x\sim \mathcal{EXD}$, the entropy condition is opposite. We choose $\overline{H_l}$ and $\overline{H_g}$ kept from training stage as the threshold and introduce the inspection index $\pi$ defined as:
\begin{equation}
\label{eq:pai}
\pi :=
  \begin{cases} 
    (H_l < \overline{H_l}) \wedge (H_g > \overline{H_g}), & \text{if}\quad\tau > 1 \\
    (H_l > \overline{H_l}) \wedge (H_g < \overline{H_g}), & \text{if}\quad\tau < 1
  \end{cases}
\end{equation}
Since the hypothesis induced by MLE is whether the new sample belongs to IND/EXD, then $\pi$ defined in \eqref{eq:pai} can be used as the condition for hypothesis testing: if $\pi=1$, then the hypothesis is accepted; Otherwise if $\pi=0$, the hypothesis is rejected .Therefore the events detection formula can be built on $\pi$ and $\tau$ as:
\begin{equation}  
\label{eq:ed}  
(\alpha, \beta) \rightarrow\left\{  
\begin{array}{ll}  
(\alpha+1, \beta), & \text{if }\pi=1\text{ and } \tau>1\\  
(\alpha, \beta+1), & \text{if }\pi=1\text{ and } \tau<1\\  
(\alpha, \beta),    & \text{if }\pi=0  
\end{array}\right.  
\end{equation}

\textbf{Control of HBU.}
It can be seen from \eqref{eq:ed} that if test samples are constantly coming, then parameters of the historical prior $Beta(m;\alpha,\beta)$ will grow so much that new timing information hardly affect the accumulated parameters. Thus, we introduce a threshold hyperparameter $\lambda$ to control the strength of prior. Concretely, we implement HBU with Algorithm ~\ref{alg:adjust_prior}, which is executed once prior update is done.
 
\begin{algorithm}
\caption{Algorithm for Pruning Historical Prior Strength}
\label{alg:adjust_prior} 
\begin{algorithmic}[1]
\State \textbf{Input:} Updated values $(\alpha, \beta)$ and threshold $\lambda$.
\If{$\alpha + \beta > \lambda$}
    \State Update $\alpha \gets 1 + \frac{\alpha}{\alpha + \beta}$
    \State Update $\beta \gets 1 + \frac{\beta}{\alpha + \beta}$
\EndIf
\State \textbf{Output:} Pruned values $(\alpha, \beta)$.
\end{algorithmic}
\end{algorithm}

\subsection{Characteristic Bayesian Update (CBU)}

\textbf{Bayesian update and challenges.}
CBU emphasizes the characteristic of the current test feature to modify the prior of $m$. Specifically, CBU operates based on the following Bayesian formula,
\begin{equation}
\mathrm{P}(x \sim \mathcal{EXD} \mid \hat{\mathbf{z}}) = 
\frac{\mathrm{P}(x \sim \mathcal{EXD}) \cdot q_{dis}(\hat{\mathbf{z}} \mid x \sim \mathcal{EXD})}{\mathrm{P}(x \sim \mathcal{EXD}) \cdot q_{dis}(\hat{\mathbf{z}} \mid x \sim \mathcal{EXD}) \\
\quad + \mathrm{P}(x \sim \mathcal{IND}) \cdot q_{dis}(\hat{\mathbf{z}} \mid x \sim \mathcal{IND})}.
\label{eq:PDB}
\end{equation}

where $q_{dis}(\hat{\mathbf{z}}|x \sim \mathcal{IND})$ and $q_{dis}(\hat{\mathbf{z}}|x \sim \mathcal{EXD})$ are the approximated likelihood of $\hat{\mathbf{z}}$ under different priors; $\mathrm{P}(x \sim \mathcal{EXD})$ is the prior; $\mathrm{P}(x\sim \mathcal{EXD} \mid \hat{\mathbf{z}})$ is the posterior and we \emph{define a new distribution} $\mathcal{B}(1,\hat{m} \mid \hat{\mathbf{z}})$ to describe it; The denominator forms the evidence since $x \sim \mathcal{IND}$ and $x \sim \mathcal{EXD}$ are opposite events. CBU derives parameter estimation of $\mathcal{B}(1,\hat{m} \mid \hat{\mathbf{z}})$ by simplifying \eqref{eq:PDB} to have
\begin{equation}
\label{eq:simple CBU}
\hat{m}:=\mathrm{P}(x\sim \mathcal{EXD} \mid \hat{\mathbf{z}}) =\frac{m}{m+(1-m)\tau}
\end{equation}
Here we face a challenge that our likelihoods attained from local and global DLEs are not accurate, so we further take advantage of entropy to rectify them. On the other hand, we find the dimensional problem mentioned in Section ~\ref{sec:motivation} again emerged, causing estimation variance positively correlated with dimension, which may turn to divergent results under a high dimension.

\textbf{Solutions.}
To solve the variance problem, we empirically add an index $\frac{1}{d}$ on output of each dimension to reduce variance to a fair scale to avoid the curse of dimensionality, which is equivalent to replacing $q_{dis}$ with the geometry average $\hat{q}_{dis}:={q_{dis}}^\frac{1}{d}$ as a result.

To further rectify $\hat{q}_{dis}$ with entropy, we consider an intuitive solution. First, \eqref{eq:DLE} shows that range of discretised likelihood is $[0,1]$, and we believe a modified likelihood still maintains this property. A natural approach is to employ a power function with exponent greater than $0$. We denote the chosen exponent as \( u \) to transform $\hat{q}$ into $\hat{q}^u$. The higher the value of \( u \), the smaller the function value. As previously discussed, entropy serves as an approximation of empirical loss; thus, the lower the entropy, the more trustworthy the model. Consequently, when an entropy criterion favors one party, we prefer a smaller value of \( u \). Moreover, considering the heterogeneous nature of data in FL, personal classifiers individually trained on client side tend to achieve lower entropy due to lower inherent complexity of their data. In contrast, the global classifier faces richer data source and is at disadvantage when comparing entropy. To balance the factors of the model's inherent capability, we introduce \eqref{eq:rel-en} to serve as the logarithm of \( u \),
\begin{equation}
\mathrm{log(u_l)} := \frac{H_l - \overline{H_l}}{\overline{H_l}}; 
\mathrm{log(u_g)} := \frac{H_g - \overline{H_g}}{\overline{H_g}} 
\label{eq:rel-en}
\end{equation}

Here the reason for using the algorithm is to ensure that \( u > 0 \).

Finally, we modify $\tau$ to $\hat{\tau}$ as the following form:
\begin{equation}
\label{eq:tau}
\hat{\tau}:=\frac{\hat{q}_{dis}(\hat{\mathbf{z}}|x \sim \mathcal{IND})^{\exp \left(\frac{H_l-\overline{H_l}}{\overline{H_l}}\right)}}{\hat{q}_{dis}(\hat{\mathbf{z}}|x \sim \mathcal{EXD})^{\exp \left(\frac{H_g-\overline{H_g}}{\overline{H_g}}\right)}} 
\end{equation}

\begin{figure*}[]
\centering
\includegraphics[width=2\columnwidth]{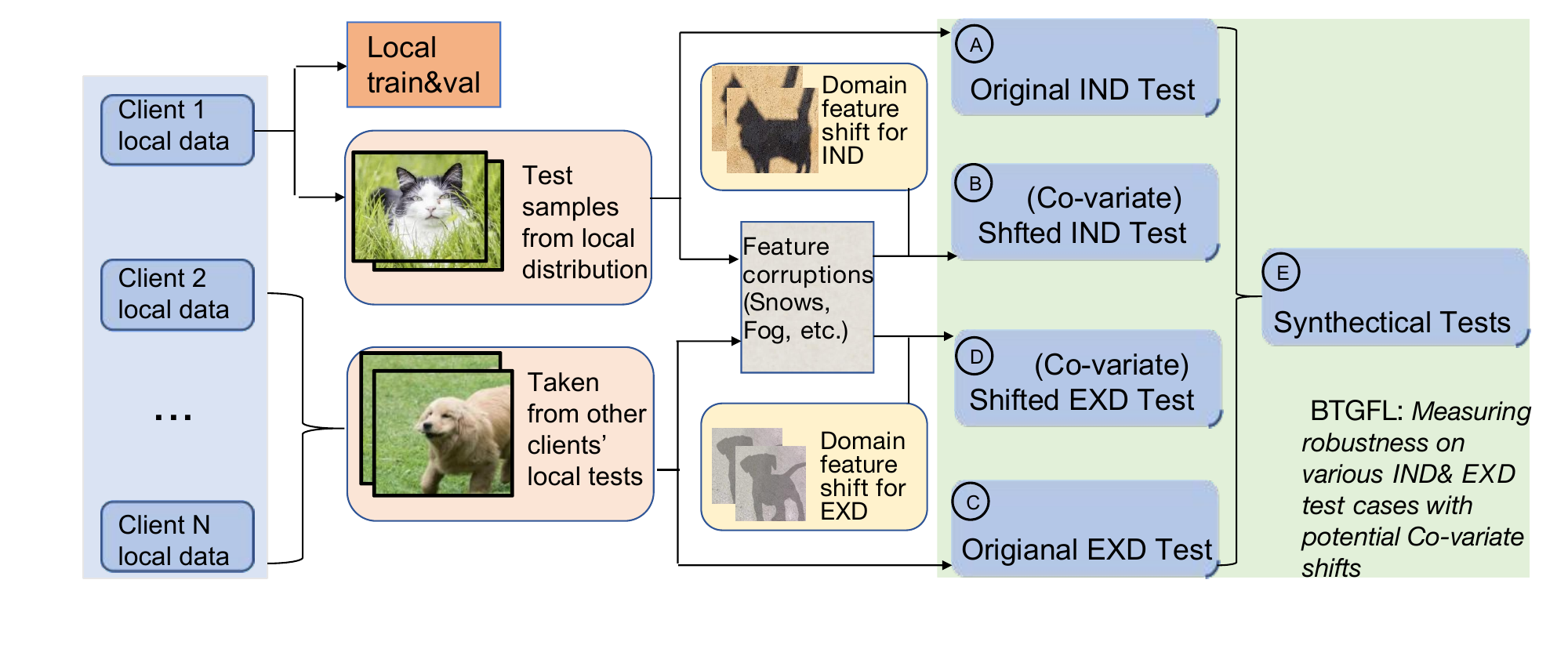} 
\vskip -0.1in
\caption{Illustration of our benchmark (BTGFL) for evaluating test-time generalization under the FL context.}
\label{fig:BTGFL}
\end{figure*}

\subsection{Dual Posterior Injection (DPI)} 
\textbf{Posterior from estimation.} HBU builds a prior $Beta(m;\alpha,\beta)$ for CBU, while CBU further updates $m$ to posterior $\hat{m}$ according to $\hat{\tau}$. Based on them, we can substitute \eqref{eq:PDB} and \eqref{eq:tau} into $Beta(m;\alpha,\beta)$ to derive the analytic expression for $pdf(\hat{m} \mid \mathbf{z})$ by the following probability distribution transformation formula of random variables.

If $X$, $Y$ are random variables and $Y = g(X)$, then
\begin{equation}
\label{eq:pdt}
pdf_Y(y) = pdf_X(g^{-1}(y)) \left| \frac{d}{dy} g^{-1}(y) \right|
\end{equation}
 where $pdf_X(x)$ and $pdf_Y(y)$ are probability density functions of $X$ and $Y$, $g^{-1}(y)$ is the inverse function of $g(y)$. 

According to \eqref{eq:pdt}, we have $\hat{m}=g(m)=\frac{m}{m+(1-m)\tau}$ and $pdf(m)=Beta(m;\alpha,\beta)$. Thus it can be deduced that 
\begin{equation}
g^{-1}(\hat{m})=\frac{\hat{m} \cdot \hat{\tau}}{1+\hat{m} \cdot \hat{\tau} - \hat{m}}
\end{equation}
and
\begin{equation}
\frac{dg^{-1}(\hat{m})}{d\hat{m}} =(\frac{\hat{m} \cdot \hat{\tau}}{1+\hat{m} \cdot \hat{\tau} - \hat{m}})'=\frac{\hat{\tau}}{\left(1+\hat{m} \cdot \hat{\tau} - \hat{m}\right)^2}>0
\end{equation}
So  the transformed $pdf(\hat{m})$ is achieved by
\begin{equation}
\label{eq:pdf}
pdf(\hat{m} \mid \mathbf{z})= Beta(\frac{\hat{m} \cdot \hat{\tau}}{1+\hat{m} \cdot \hat{\tau} - \hat{m}};\alpha,\beta) \cdot \frac{\hat{\tau}}{\left(1+\hat{m} \cdot \hat{\tau} - \hat{m}\right)^2}
\end{equation}
\textbf{Posterior from two heads.} Recall the probability equation  
\begin{equation}
\label{eq:result}
psi(\mathbf{y} \mid \mathbf{z})=\int  psi(\mathbf{y}  \mid \hat{m},\mathbf{z}) \cdot pdf(\hat{m}\mid \mathbf{z})\, d\hat{m}
\end{equation}
given in Section ~\ref{sec:motivation}, now we can substitute $pdf(\hat{m} \mid \mathbf{z})$ derived from \eqref{eq:pdf} and another posterior $psi(\mathbf{y}  \mid \hat{m})$ derived from \eqref{eq:psi} into \eqref{eq:result} to obtain the \emph{final interpolated prediction result} $Y_{int}$ as follows.

\begin{equation}
\label{eq:finish}
Y_{int}:=psi(\mathbf{y} \mid \mathbf{z})= e \cdot Y_g +(1-e) \cdot Y_l,
\end{equation}
\[
\text{where } e = \int \hat{m} \cdot pdf(\hat{m}\mid \mathbf{z})\, d\hat{m},
\]

Through posterior injection achieved by above formulas, we deal with the historical information, current sample characteristic and knowledge contained in each head under an integral equation in \eqref{eq:finish}, which can be work out by numerical methods, to achieve the probabilistic coefficient $e$ for rational interpolated prediction.

\section{Evaluation}
\subsection{Experimental Settings}
\label{sec:ES}
\textbf{Benchmark.}
We notice that BRFL, a recent benchmark for test-time robustness of FL proposed by FedTHE~\cite{jiang2022test}, simply puts external distribution on the equal position as other kinds of co-variate shift, which does not well match the requirements of generalization in the FL field. Therefore, To fairly evaluate our method under TGFL, we propose a new \textbf{B}enchmark for \textbf{TGFL} (\textbf{BTGFL}), which considers scenarios  made up of internal and external distributions together with feature-level co-variate shift. 

Specifically, we construct 5 distinct test distributions, which are defined as: 
\textcircled{A} Original Internal test,
\textcircled{B} (Co-variate) Shifted Internal test,
\textcircled{C} Original External test,
\textcircled{D} (Co-variate) Shifted External test, and
\textcircled{E} Synthetical tests. The overall illustration of our benchmark can be described as Figure~\ref{fig:BTGFL}.

In Figure~\ref{fig:BTGFL}, \textcircled{A} reflects test set iid with local training set, aiming to investigate the personalization ability; \textcircled{B} represents co-variate shifted samples conforming same class distribution as \textcircled{A}, while half of which are domain shifted and others are corrupted samples, we mix 2 kinds of co-variate shifts together to simulate realistic scenarios; \textcircled{C} examines External distribution by drawing samples from the test data of other clients, intending to verify generalization; \textcircled{D} stands for the External counterpart of \textcircled{B}; We build \textcircled{E} with evenly selected samples from the previous 4 test distributions, this set is designed to test the Extreme Cases as it is not inclined to Internal/External or other kinds of feature Distribution, thus methods should carefully use historical information during test to avoid potential negative effects. 

Compared to BRFL, BTGFL places greater emphasis on the trade-off between internal and external distributions, which facilitates a fair evaluation on the generalization ability of different methods.
\begin{figure*}[]
\centering
\includegraphics[width=2\columnwidth]{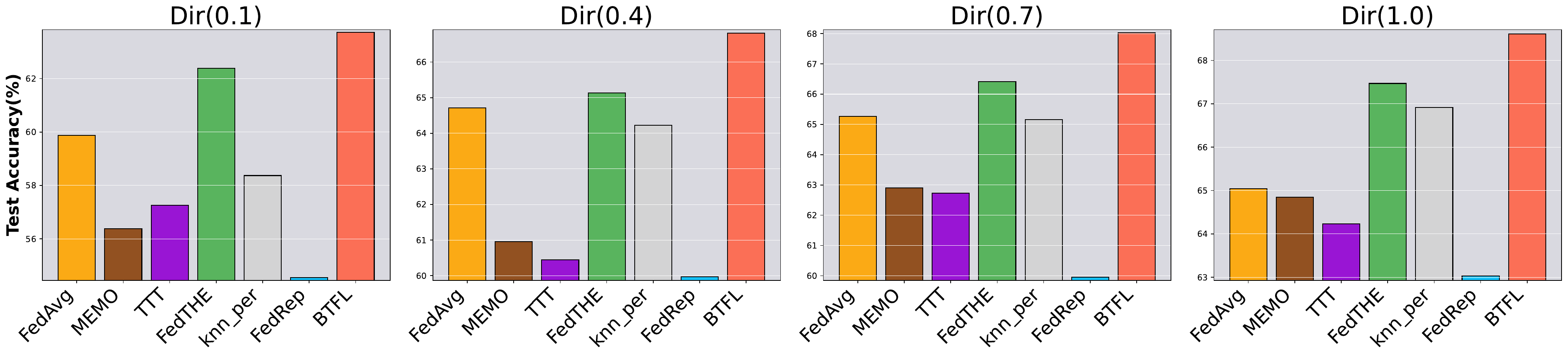} 
\vskip -0.1in
\caption{
		\textbf{Test accuracy of baselines on Synthetical tests with different degrees of heterogeneity.}
		A simple CNN is trained on CIFAR10. The client heterogeneity is determined by the value of Dirichlet distribution~\cite{yurochkin2019bayesian,hsu2019measuring}, termed as 'Dir'.
		\looseness=-1
	}
\vskip -0.1in
\label{fig:cnncifar}
\end{figure*}

\textbf{Datasets.} 
We consider OfficeHome~\cite{venkateswara2017deep}, CIFAR10~\cite{krizhevsky2009learning} and ImageNet ~\cite{deng2009imagenet} (downsampled to the resolution of 32, similar to our baselines~\cite{chrabaszcz2017downsampled}). For each dataset, following CIFAR10-C~\cite{hendrycks2018benchmarking}, we construct the corruption co-variate shifts in \textcircled{B} and \textcircled{D} by leveraging 15 common corruptions from original samples; and split CIFAR10.1 ~\cite{recht2018cifar} as domain co-variate shifts for CIFAR10 in \textcircled{B} and \textcircled{D}; For ImageNet32, domain shifted samples are built by splitting ImageNet-R~\cite{hendrycks2021many} to each client based on its local class distribution; Since OfficeHome is already separated as 4 various domains, we leave a single domain as domain shifted samples and other 3 domains are treated as In-distribution. Train and Test batch size is separately set to 128 for experiments on ImageNet, and 32 for that on CIFAR10/OfficeHome if not otherwise specified.

\textbf{Models.}
On the CIFAR-10 dataset, we conducted experiments on simple CNN architecture~\cite{lecun1998gradient,chen2022on}(w/o BN layers), ResNet20~\cite{he2016deep}(w/ GN~\cite{hsieh2020non}), and 4-layer Compact Convolutional Transformer (CCT4)~\cite{hassani2021escaping}. On ImageNet and OfficeHome, we performed experiments on ResNet20-GN. For more details please refer to Appendix \ref{sec:expdetial}.

\textbf{Baselines.}
In the realm of FL, various strong baselines are considered from different paradigms such as FL, personalized FL (PFL), and test-time adaptation (TTA), which can be easily applied to FL contexts:
\begin{itemize}
\item \textbf{FedAvg}: As a foundational approach, \texttt{FedAvg}~\cite{mcmahan2017communication} is designed to learn a global model that is shared across all clients. 
\item \textbf{TTA methods plus FedAvg}: These methods are adapted for personalized FL by augmenting the \texttt{FedAvg + FT} personalized model with test-time adaptation. This means that prediction for each test point is performed immediately after the model is adapted to the local data. The TTA methods considered include:
   \texttt{TTT} (online version)~\cite{sun2020test},
   \texttt{MEMO}~\cite{zhang2021memo}.
\item \textbf{FedTHE}~\cite{jiang2022test}: The most competitive baseline designed for the robustness of test-time FL. 
\end{itemize}

Besides, we omit the comparison with ATP~\cite{bao2024adaptive} and FedICON~\cite{tan2023taming}. ATP argues attention to the performance on novel clients, which diverges from the scenario of TGFL; FedICON focuses on the test-time intra-client heterogeneity belonging to IND shifts, and ignores the adaptation to EXD shifts, which is also not applicable to our TGFL settings. Here we fix the testing batch size to 32 for all methods since they are robust to the batch size except for FedTHE, whose limitation has been illustrated in Table \ref{tab:EMA}.

\begin{table}[!h]

    \centering
    \caption{
		\textbf{Test accuracy of baselines across various test distributions under Dir(0.1).} A simple CNN is trained on CIFAR10. 
		\looseness=-1
	}
 \label{tab:cifar10}
    \resizebox{.48\textwidth}{!}{%
        \begin{tabular}{l|ccccc|c}
			\toprule
			Methods           & \parbox{2.cm}{ \centering Original IND} & Shifted IND                  & Original EXD                  & Shifted EXD                 & \parbox{1.5cm}{ \centering    Synthetical                           } & \parbox{1.5cm}{ \centering Avg}    \\ \midrule
            FedAvg & 65.60 $_{\pm 0.24}$ & 51.91 $_{\pm 0.07}$ & 65.21 $_{\pm 0.25}$ & 51.77 $_{\pm 0.07}$ & 58.88 $_{\pm 0.13}$ & 58.67 \\
        
            FedAvg+FT & 88.18 $_{\pm 0.02}$ & 81.64 $_{\pm 0.06}$ & 31.72 $_{\pm 0.05}$ & 26.06 $_{\pm 0.20}$ & 56.78 $_{\pm 0.13}$ & 56.88 \\
            
            FedRep                                  & 88.98 $_{\pm 0.31}$                                   & 82.19 $_{\pm 0.27}$                                   & 25.84 $_{\pm 0.00}$          & 22.63 $_{\pm 0.08}$          & 54.71 $_{\pm 0.05}$    & 54.87 \\
            FedRoD                                  & 89.13 $_{\pm 0.16}$                                   & 82.67 $_{\pm 0.10}$                                   & 26.80 $_{\pm 0.10}$          & 22.85 $_{\pm 0.04}$          & 55.33 $_{\pm 0.36}$    & 55.36\\
            kNN-Per                                 & 89.63 $_{\pm 0.69}$                                   & 82.00 $_{\pm 0.61}$                                   & 34.73 $_{\pm 3.45}$          & 27.13 $_{\pm 1.53}$          & 58.37 $_{\pm 0.77}$    & 58.37\\
            \midrule
             FedTHE & 87.16 $_{\pm 0.05}$ & 74.95 $_{\pm 0.44}$ & 59.93 $_{\pm 0.33}$ &  46.32 $_{\pm 1.04}$ & 62.38 $_{\pm 0.85}$ & 66.15 \\
            TTT                                     & 88.77 $_{\pm 0.01}$                                   & 82.33 $_{\pm 0.03}$                                   & 32.69 $_{\pm 0.07}$          & 26.63 $_{\pm 0.10}$          & 57.15 $_{\pm 0.02}$    & 57.51\\
            MEMO                                    & 89.24 $_{\pm 0.21}$                                   & 82.77 $_{\pm 0.16}$                                   & 28.16 $_{\pm 0.06}$          & 23.13 $_{\pm 0.05}$          & 55.97 $_{\pm 0.22}$    & 55.85\\

            \midrule
            BTFL (Ours) & 88.12 $_{\pm 0.11}$ & 79.74 $_{\pm 0.05}$ & 60.23 $_{\pm 0.75}$ & 48.31 $_{\pm 0.14}$ & 63.73 $_{\pm 0.41}$ &  \textbf{68.03} \\
            \bottomrule
        \end{tabular}%
    }
\end{table}

\subsection{Experimental Results}

\textbf{Superior IND\&EXD accuracy of our approaches across diverse test scenarios and FL settings.} Table \ref{tab:cifar10} exhibits accuracy performance of BTFL under Dir(0.1). Besides, We conducted extensive baseline comparisons for BTFL in Figure \ref{fig:cnncifar}, including different levels of non-independent and identically distributed (non-iid) data. It is interesting to find that the intuition-driven method FedTHE, with limitations mentioned in Section \ref{sec:motivation}, achieves comparable accuracy with the theory-guided BTFL. However, recall that both BTFL and FedTHE are test-time methods that just extract knowledge from a known model instead of acquiring new knowledge from test data flow, thus the upper promotion bound is constrained by the knowledge implied in parts of a frozen model, which accounts for the 1-2\% advantage to FedTHE made by BTFL.

\begin{figure}[!htb]
  \begin{minipage}{.45\textwidth}
    \hspace*{-0.6cm} 
    \includegraphics[width=1.2\textwidth]{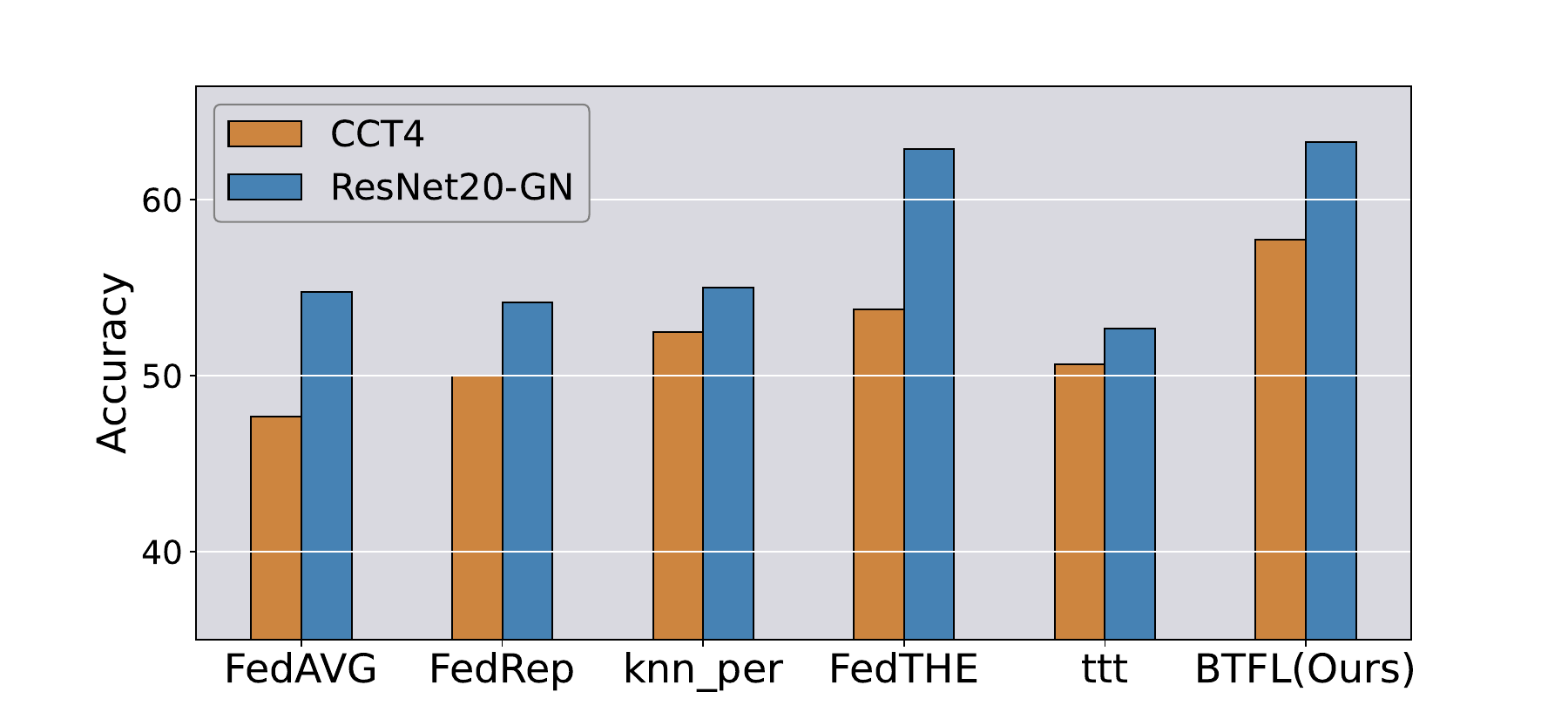} 
  \end{minipage}\hfill
  \begin{minipage}{.45\textwidth}
    \centering
    \caption{The test accuracy of benchmark models (encompassing the training of both ResNet20-GN and CCT4 on the CIFAR10 dataset with a heterogeneity factor of Dir(0.1)) are presented. Our methods are justified against five strong competitors, with the outcomes averaged across 5  IND/EXD test sets in BTGFL. }
    \label{fig:cifar_cct_res}
  \end{minipage}
\end{figure}

\textbf{Applicability to various datasets\&models\&test shifts.} 
Table \ref{tab:examining_one_arch_on_different_imagenet_variants_with_fixed_non_iid_ness}and \ref{tab:examining_one_arch_on_different_OfficeHome_variants_with_fixed_non_iid_ness}  look at the challenging OfficeHome and ImageNet dataset. Our method provides consistent accuracy of IND tests with TTA methods while  maintains the improved EXD accuracy across different competitors. It is worth noting that under the Synthetical test, all methods face significant accuracy drop, remaining lots of room for optimization for testing methods when facing practical scenarios. Compared with Cifar10, BTFL stands out by a larger margin to demonstrate the scalability of our method. Besides, the benefits of our approaches are further pronounced in Figure \ref{fig:cifar_cct_res}, which summarizes the accuracy of strong baselines on SOTA models (i.e. ResNet20 with GN ~\cite{hsieh2020non} and CCT4 with LN ~\cite{hassani2021escaping}). 

\textbf{Efficiency for Inference.}
As a test-time method, the proposed analytical method BTFL requires little computation and negligible extra inference usage of GPU memory, which enables convenient deployment on edge devices under the constrained resources. We record the time cost to infer 1,000 samples for different methods in Table \ref{tab:speed} to quantify our inference efficiency. We can clearly see that BTFL can accelerate the adaptation process by roughly 4X-12X compared to other baselines, plus with advanced accuracy performance. Our method achieves similar speed under different model structures, this relies on  BTFL's property of dispatching from model architecture, which makes it resilient to complexity of current large models.

\begin{table}[!h]
	\caption{
		\textbf{Computational time (s) of various TTA methods for 1,000 CIFAR-10 samples under various model architectures.} Test batch is set to 1 for all methods. Time in seconds recorded in table is averaged over the results from 20 clients' local tests.
		\looseness=-1
	}
	\label{tab:speed}
	\centering
	\begin{tabular}{l|cccc}
		\toprule
		  & TTT & MEMO & FedTHE & BTFL(Ours) \\ \midrule
		CNN  & 53.96 $_{\pm 0.67}$ & 160.99 $_{\pm 3.16}$ & 77.72 $_{\pm 0.18}$ & \textbf{17.30 $_{\pm 0.12}$} \\
        ResNet20  & 78.86 $_{\pm 0.89}$ & 212.48 $_{\pm 4.23}$ & 69.22 $_{\pm 0.36}$ & \textbf{17.94 $_{\pm 0.19}$} \\
        CCT  & 119.18 $_{\pm 1.45}$ & 222.00 $_{\pm 5.22}$ & 83.03 $_{\pm 0.84}$ & \textbf{18.15$_{\pm 0.21}$} \\
		\bottomrule
	\end{tabular}
\end{table}

\begin{table}[!h]
	\caption{
		\textbf{Test accuracy of baselines on different ImageNet test distributions.}
		ResNet20-GN with width factor of $4$ is used for training ImageNet on clients with heterogeneity Dir(0.1), and results on Dir(1.0) can be found in Appendix \ref{sec:iidexp}.
		\looseness=-1
	}
	\label{tab:examining_one_arch_on_different_imagenet_variants_with_fixed_non_iid_ness}
	\centering
	\resizebox{.48\textwidth}{!}{%
		\begin{tabular}{l|ccccc|c}
			\toprule
			Methods           & \parbox{2.cm}{ \centering Original IND} & Shifted IND                  & Original EXD                  & Shifted EXD                 & \parbox{1.5cm}{ \centering    Synthetical                           } & \parbox{1.5cm}{ \centering Avg}    \\ \midrule
            FedAvg            & 62.10 $_{\pm 0.11}$                               & 19.31 $_{\pm 0.14}$         & 62.04 $_{\pm 0.20}$          & 19.13 $_{\pm 0.25}$          & 40.72 $_{\pm 0.01}$                & 40.66           \\      
            FedAvg+FT         & 78.81 $_{\pm 0.03}$                               & 34.04 $_{\pm 0.13}$         & 19.02 $_{\pm 0.02}$          & 6.50 $_{\pm 0.02}$           & 34.47 $_{\pm 0.04}$                & 34.57       \\
			FedRoD            & 80.31 $_{\pm 0.15}$                               & 35.21 $_{\pm 0.60}$         & 17.07 $_{\pm 0.14}$          & 6.33 $_{\pm 0.02}$           & 34.72 $_{\pm 0.13}$                & 34.73        \\
			FedRep            & 76.38 $_{\pm 0.55}$                               & 33.45 $_{\pm 0.54}$         & 13.42 $_{\pm 0.16}$          & 5.25 $_{\pm 0.01}$           & 32.07 $_{\pm 0.05}$                & 32.11       \\
            kNN-Per           & 83.81 $_{\pm 0.51}$                               & 35.56 $_{\pm 0.20}$         & 42.68 $_{\pm 0.59}$          & 13.57 $_{\pm 0.11}$          & 43.93 $_{\pm 0.40}$                & 43.91        \\
            \midrule
             FedTHE            & 88.09 $_{\pm 0.36}$                               & 33.65 $_{\pm 0.95}$         & 50.57 $_{\pm 2.97}$          & 15.03 $_{\pm 2.63}$          & 43.41 $_{\pm 0.75}$                & 46.15       \\  
            MEMO              & 75.71 $_{\pm 0.87}$                               & 27.66 $_{\pm 0.45}$         & 15.77 $_{\pm 0.40}$          & 4.33 $_{\pm 3.25}$           & 29.62 $_{\pm 1.13}$                & 30.62       \\
			TTT               & 79.19 $_{\pm 0.03}$                               & 39.73 $_{\pm 0.02}$         & 16.49 $_{\pm 0.04}$          & 5.87 $_{\pm 0.00}$           & 33.83 $_{\pm 0.02}$                & 35.02       \\    
            \midrule
			BTFL(Ours)        & 87.01 $_{\pm 0.25}$                               & 32.74 $_{\pm 0.15}$          & 60.94 $_{\pm 0.07}$         & 18.21 $_{\pm 0.05}$          & 45.41 $_{\pm 0.01}$               & \textbf{48.86}         \\
			\bottomrule
		\end{tabular}%
	}
\end{table}

\begin{table}[!h]
	\caption{
		\textbf{Test accuracy of baselines on different OfficeHome test distributions.}
		ResNet20-GN with width factor of $4$ is used for training OfficeHome on clients with heterogeneity Dir(0.1), and results on Dir(1.0) can be found in Appendix \ref{sec:iidexp}.
		\looseness=-1
	}	\label{tab:examining_one_arch_on_different_OfficeHome_variants_with_fixed_non_iid_ness}
	\centering
	\resizebox{.48\textwidth}{!}{%
		\begin{tabular}{l|ccccc|c}
			\toprule
			Methods           & \parbox{2.cm}{ \centering Original IND} & Shifted IND                  & Original EXD                  & Shifted EXD                 & \parbox{1.5cm}{ \centering    Synthetical                           } & \parbox{1.5cm}{ \centering Avg}    \\ \midrule
            FedAvg            & 36.84 $_{\pm 0.37}$                               &  21.55$_{\pm 0.21}$           &36.86 $_{\pm 1.31}$ &24.78 $_{\pm 0.50}$ &  30.11$_{\pm1.01 }$                &30.03           \\      
            FedAvg+FT         & 51.56 $_{\pm 1.93}$                               & 34.31 $_{\pm 0.67}$          & 6.30 $_{\pm 0.05}$          & 6.58 $_{\pm 0.06}$          & 24.67 $_{\pm 0.76}$                & 24.68          \\
			FedRoD            & 45.94 $_{\pm 2.37}$                               & 28.42 $_{\pm 1.35}$          & 5.93 $_{\pm 0.41}$          & 5.88 $_{\pm 0.12}$          & 21.60 $_{\pm 1.41}$                & 21.55      \\
			FedRep            & 53.83 $_{\pm 3.09}$                               & 34.99 $_{\pm 1.08}$          & 7.38 $_{\pm 0.20}$          & 7.36 $_{\pm 0.19}$          & 25.65 $_{\pm 1.22}$                & 25.84         \\     
            kNN-Per           & 60.65 $_{\pm 0.51}$                               & 38.61 $_{\pm 0.45}$          & 21.23 $_{\pm 0.36}$         & 16.09 $_{\pm 2.00}$         & 33.88 $_{\pm 1.31}$                & 34.09        \\
            \midrule
             FedTHE            & 67.38 $_{\pm 1.16}$                      & 42.41 $_{\pm 0.69}$ & 11.30 $_{\pm 0.79}$          & 10.07 $_{\pm 0.38}$        & 32.44 $_{\pm 0.98}$                & 32.72        \\   
             MEMO           &   29.77 $_{\pm0.99 }$                   &  19.95   $_{\pm 1.23}$ &  3.01$_{\pm 0.39}$          &  3.80$_{\pm 0.74}$        & 14.28 $_{\pm 0.65}$                &    14.16    \\   
			TTT               & 54.90 $_{\pm 1.06}$                               & 35.26 $_{\pm 0.65}$          & 6.06 $_{\pm 0.33}$          & 6.23 $_{\pm 2.64}$          & 25.16 $_{\pm 0.67}$                & 25.52        \\
              
            \midrule
			BTFL(Ours)        & 65.21 $_{\pm 1.00}$                               & 39.08 $_{\pm 0.45}$          & 28.05 $_{\pm 4.13}$          & 19.80 $_{\pm 1.23}$         & 34.07 $_{\pm 1.92}$               & \textbf{37.24}        \\
			\bottomrule
		\end{tabular}%
	}
\end{table}

\section{Conclusion}
In this paper, we introduce Test-time Generalization for Internal and External Distributions in Federated Learning (TGFL), emphasizing the need for models to adapt under IND/EXD shifts during testing. Our proposed method, BTFL, uses a Bayesian approach to balance personalization and generalization at the test time. BTFL features a dual-headed architecture that leverages both local and global knowledge, making interpolated predictions by blending historical and current information under a novel probability perspective. Experiments show BTFL improves performance across different datasets, FL settings and models, with low time consumption.

\section*{Acknowledgments}
We would like to thank the anonymous reviewers for their valuable feedback. This work was partly supported by the National Natural Science Foundation of China (62302054) and CCF-Huawei Populus Euphratica Fund (System Software Track, CCF-HuaweiSY202410).



\appendix
\section{Mathematical proof of formulas in the text}

\subsection{Beta-Bernoulli process}
\label{sec:bbp}
The beta distribution describes a continuous random variable with a domain between 0 and 1, which is the range of values for the probability of an event. The definition of its probability density function is
\begin{equation}
\begin{split}
\operatorname{Beta}(\theta ; \alpha, \beta) &:= \frac{\theta^{\alpha-1} (1-\theta)^{\beta-1}}{\int_{0}^{1} u^{\alpha-1} (1-u)^{\beta-1} \, d u} \\
&= \frac{\Gamma(\alpha+\beta)}{\Gamma(\alpha) \Gamma(\beta)} \theta^{\alpha-1} (1-\theta)^{\beta-1} \\
&= \frac{\theta^{\alpha-1} (1-\theta)^{\beta-1}}{B(\alpha, \beta)} \quad (0 < \theta < 1)
\end{split}
\end{equation}
where
\begin{equation}
\Gamma(\mathrm{x}):= \int_{0}^{+\infty} t^{x-1} * e^{-t} d t 
\end{equation}
and 
\begin{equation}
B(\alpha, \beta):= \frac{\Gamma(\alpha+\beta)}{\Gamma(\alpha) * \Gamma(\beta)} 
\end{equation}
While the probability distribution of a binomial distribution is:
\begin{equation}
q(X=k;\theta,n) = \binom{n}{k} \theta^k (1-\theta)^{n-k}
\end{equation}

Because the definition domain of the beta distribution is $[0,1]$, beta distribution can be chosen as the estimate of the binomial distribution parameter. So when we observe x positive events in n binomial experiments, we can use Bayesian formula to update the prior distribution.

By the complete proof stated in \eqref{eq:bbp}, it can be concluded that the updated distribution is still a beta distribution. This property is also known as conjugate distribution:
\begin{equation}
\label{eq:bbp}
\begin{split}
\mathrm{p}(\theta \mid \mathrm{k}) &= \frac{\operatorname{Beta}(\theta ; \alpha, \beta) * \mathrm{q}(\mathrm{k} \mid \theta)}{\mathrm{P}(\mathrm{k})} \\
&= \frac{\operatorname{Beta}(\theta ; \alpha, \beta) * \mathrm{P}(\mathrm{k} \mid \theta)}{\int \operatorname{Beta}(\vartheta ; \alpha, \beta) * \mathrm{P}(\mathrm{k} \mid \vartheta) \mathrm{d} \vartheta} \\
&= \frac{\frac{\theta^{\alpha-1} *(1-\theta)^{\beta-1}}{B(\alpha, \beta)} * \binom{n}{k} \theta^{k}(1-\theta)^{n-k}}{\int \frac{\vartheta^{\alpha-1} *(1-\vartheta)^{\beta-1}}{B(\alpha, \beta)} * \binom{n}{k} \vartheta^{k}(1-\vartheta)^{n-k} \mathrm{d} \vartheta} \\
&= \frac{\theta^{\alpha+k-1} *(1-\theta)^{\beta+n-k-1}}{\int \vartheta^{\alpha+k-1} *(1-\vartheta)^{\beta+n-k-1} \mathrm{d} \vartheta} \\
&= \frac{\theta^{\alpha+k-1} *(1-\theta)^{\beta+n-k-1}}{B(\alpha+k, \beta+n-k)} \\
&= \operatorname{Beta}(\alpha+k, \beta+n-k) 
\end{split}
\end{equation}
The posterior is $\operatorname{Beta}(\alpha+k, \beta+n-k)$ after new events emerges. \eqref{eq:bbp} is also referred as Beta-Bernoulli process~\cite{staton2018beta}.

On the other hand, our prior parameter for the beta distribution before HBU is initialized to (1,1), corresponding to the probability density function:
\begin{equation}
\operatorname{Beta}(m ; 1,1)=\frac{m^{0} *(1-m)^{0}}{B(1,1)}=1
\end{equation}
Thus it conforms to $\mathcal{U}(0,1)$, the uniform distribution on the interval [0,1], which well satisfies the principle of insufficient reasons~\cite{dubs1942principle} and is also suitable for the Beta-Bernoulli process.

We additionally give an explanation about the principle of insufficient reasons: In the absence of any specific rationale or information to suggest that one event is more likely to occur than another, we should presume that all possible events occur with equal probability.

\subsection{Entropy as lower-bound of empirical risk loss}
\label{sec:lb}
In general, the loss function of a deep learning model, known as Cross-Entropy, compels the model to produce an output distribution that closely approximates a one-hot encoding,
\begin{equation}
\mathrm{L}(\mathrm{x})=\text { Cross-Entropy }\left(\mathrm{Y}_{\text {true}}, \mathrm{Y}_{\text {pred}}\right) 
\end{equation}
By definition, we have:
\begin{equation}
\text{Cross-Entropy}\left(\mathrm{Y}_{\text{true}}, \mathrm{Y}_{\text{pred}}\right) = \mathrm{H}\left(\mathrm{Y}_{\text{pred}}\right) + \mathrm{D}_{\mathrm{KL}}\left(\mathrm{Y}_{\text{true}} \parallel  \mathrm{Y}_{\text{pred}}\right) 
\end{equation}
The KL divergence, denoted as \( D_{KL} \), measures the difference between two probability distributions \( P \) and \( Q \) and is defined as:
\begin{equation}
D_{KL}(P \parallel Q) := \sum_{x} P(x) \log\left(\frac{P(x)}{Q(x)}\right)>0
\end{equation}
The objective of training is to minimize the empirical risk:
\begin{equation}
\min\quad\mathrm{L}(\mathrm{x}) 
\end{equation}
Combining the four formulas above, it can be easily deduced that entropy is a lower bound and a fair approximation of $\mathrm{L}(\mathrm{x})$ under unsupervised conditions.

\begin{figure}[]
\centering
\includegraphics[width=1\columnwidth]{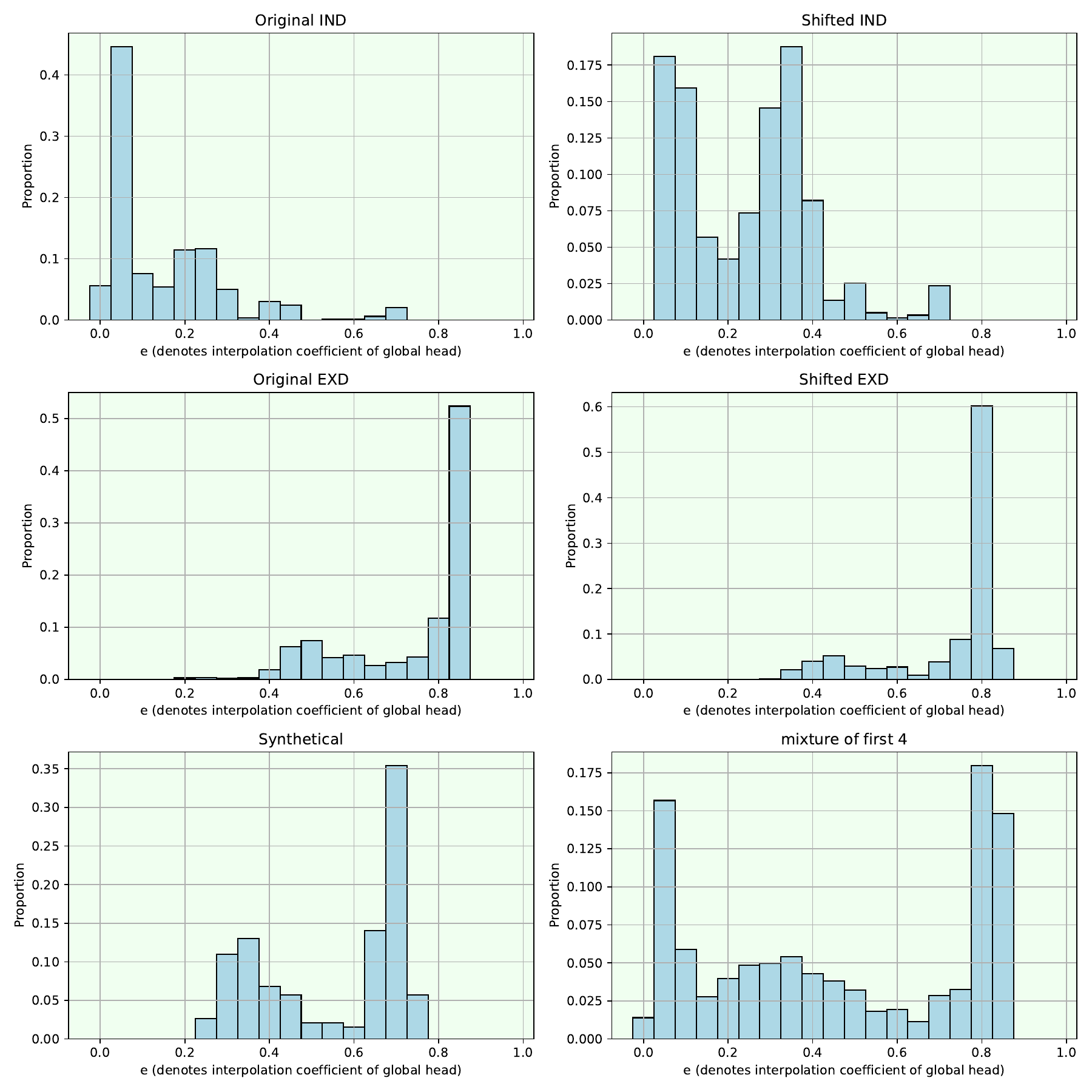} 
\vskip -0.1in
\caption{The distribution of $e$ on multiple distributions in BTGFL.}
\vskip -0.1in
\label{fig:ensemble}
\end{figure}

\section{Implementation details}
\label{sec:expdetial}
All our experiments are simulated and conducted in a server that has 3 GeForce GTX 3090 GPUs, 48 Intel Xeon CPUs, and 128GB memory. 

\textbf{Training settings.} We involve 20 fully participating clients and 130 rounds for training on CIFAR10; 10 clients and 100 rounds on ImageNet-32; 10 clients and 50 rounds on OfficeHome. Each round consists of a 5-epoch local training followed by a 1-epoch personalization phase, with only local training parameters being uploaded to the server. When training CNNs on CIFAR10, we employ an SGD optimizer at an initial learning rate of 0.01, with a weight decay set to 5e-4; For ResNet20-GN, trained on CIFAR10, OfficeHome and ImageNet-32, we follow a similar setup to the CNN on CIFAR10 but use an SGDm optimizer with a momentum factor of 0.9. In ResNet20-GN, we also set the GroupNorm~\cite{wu2018group}  to have 2 groups; For CCT4 training on CIFAR10, we opt for the Adam~\cite{kingma2014adam} optimizer, starting with an initial learning rate of 0.001 and using the default coefficients for the first and second moments (0.9 and 0.999 respectively).

\textbf{Hyperparameters.} We clarify hyperparameters used for baseline methods along with BTFL.
\begin{itemize}
\item For kNN-Per, we adopt the configuration of using $10$ neighbors and a scaling factor of $1.0$. We meticulously fine-tune the critical hyperparameter with a 0.1 step size within the range [0.0, 0.9] for each client to optimize performance on in-distribution data.
\item For FedTHE, we follow its original setup to keep both $\alpha$ and $\beta$ $0.1$; the optimization steps for solving ~\eqref{eq:slw} is set to 20.
\item For TTT,  we apply 0.001 learning rate and 1 optimization step which is the optimal.
\item For MEMO, we choose among 16, 32, or 48 augmentations and 1 to 4 optimization steps. Our best results come from using 32 augmentations with 3 optimization steps at a learning rate of 0.0005.
\item For BTFL, the only method-specific hyperparameter $\lambda$ is set to 16. We conducted various tests on the variable $\lambda$ within the range $[5, 20]$, and the experimental data showed minimal variance.
\end{itemize}

\textbf{Models.} We adopt the final fully connected (FC) layer as a universal global head and an individualized local head for CNN, ResNet20-GN, and CCT4. The neural structure of CNN encompasses 2 Conv layers (boasting 32, 64 channels and a kernel size of 5) and 2 FC layers (each with 64 neurons in the hidden layers). For ResNet20, we assign a scaling factor of width to be 1, 4 and 4 when training on CIFAR10, OfficeHome and ImageNet-32, respectively. For CCT4 on CIFAR10, we adopt the exact architecture as presented in ~\cite{hassani2021escaping}. For CCT-4/3x2, complete with a learnable positional embedding. The dimension of feature representation (that is, the output dimension of the feature extractor) of CNN, ResNet20 and CCT4 on CIFAR10 are 64, 64 and 128; OfficeHome and ImageNet-32 on ResNet20 are 512 and 256, respectively. By the way, we place $relu$ before classification head of each model to ensure effectiveness of \eqref{eq:FSQ}.

\section{More Experiment results}
\subsection{Various degrees of non-i.i.d local distributions.}

\label{sec:iidexp}
We also provide results for training ResNet20-GN on OfficeHome (see in Table \ref{tab:floff}) and downsampled ImageNet-32(see Table  \ref{tab:image1}) on Dir(1.0). Similar to the Dir(0.1) case (Table \ref{tab:examining_one_arch_on_different_OfficeHome_variants_with_fixed_non_iid_ness} and Table \ref{tab:examining_one_arch_on_different_imagenet_variants_with_fixed_non_iid_ness}),  performance advantage of our method is still
maintained.

\begin{table}[!h]
	\caption{\small
		\textbf{Test accuracy of baselines on different OfficeHome test distributions.}
		ResNet20-GN with width factor of $4$ is used for training OfficeHome on clients with heterogeneity Dir(1.0). 
		\looseness=-1
	}
	\vspace{-0.1em}
	\label{tab:floff}
	\centering
	\resizebox{.45\textwidth}{!}{%
		\begin{tabular}{l|ccccc|c}
			\toprule
			Methods           & \parbox{2.cm}{ \centering Original IND} & Shifted IND                  & Original EXD                  & Shifted EXD                 & \parbox{1.5cm}{ \centering    synthetical                           } & \parbox{1.5cm}{ \centering Average}    \\ \midrule
            FedAvg            &  50.53$_{\pm 0.96}$                               & 27.08$_{\pm 0.19}$          & 50.03$_{\pm 0.40}$         & 31.72$_{\pm 0.23}$          &  39.88$_{\pm 0.97}$                &  39.85        \\      
            FedAvg+FT         &  39.13$_{\pm 0.67}$                       & 23.23 $_{\pm 0.34}$          & 21.03 $_{\pm 1.40}$          &  14.70$_{\pm 0.43}$          &  24.34$_{\pm 1.66}$                &  24.49         \\
			FedRoD            & 40.34  $_{\pm 1.80}$                      & 22.34  $_{\pm 0.47}$          &20.18  $_{\pm 1.38}$          & 13.88$_{\pm 0.24}$          & 24.20 $_{\pm 1.11}$                & 24.19      \\
			FedRep            &  41.90$_{\pm 0.34 }$                      &  24.26 $_{\pm 0.17}$          &23.77 $_{\pm 0.17}$          & 16.22 $_{\pm 0.24}$          &  26.52$_{\pm 0.45}$                &   26.53       \\
            kNN-Per           &  56.55$_{\pm 0.81}$                               & 30.62 $_{\pm 0.73}$          & 46.86 $_{\pm 2.49}$         &  29.84$_{\pm 2.14}$         &  40.67$_{\pm 2.66}$                &  40.91       \\
			TTT               &  41.29$_{\pm 6.57}$                               &  24.54$_{\pm 2.24}$          &  24.62$_{\pm 1.61}$          &  16.43$_{\pm 0.53}$          & 26.64 $_{\pm 3.40}$                &  26.70       \\
            FedTHE            &  63.70$_{\pm4.09 }$                               & 34.80$_{\pm 0.74}$          & 37.72$_{\pm 5.36}$          &  24.93$_{\pm 2.02}$        &  40.03$_{\pm 3.92}$                &  40.23       \\      
            \midrule
			BTFL(Ours)        &  61.78$_{\pm 2.98}$                               & 33.00 $_{\pm 0.50}$          &  50.30$_{\pm 5.19}$          &  31.10$_{\pm 2.64}$         & 43.30 $_{\pm 2.47}$               & \textbf{43.90}        \\
			\bottomrule
		\end{tabular}%
	}
\end{table}

\begin{table}[!h]
	\caption{\small
		\textbf{Test accuracy of baselines on different ImageNet test distributions.}
		ResNet20-GN with width factor of $4$ is used for training ImageNet on clients with heterogeneity Dir(1.0).
		\looseness=-1
	}
	\vspace{-0.1em}
	\label{tab:image1}
	\centering
	\resizebox{.45\textwidth}{!}{%
		\begin{tabular}{l|ccccc|c}
			\toprule
			Methods           & \parbox{2.cm}{ \centering Original IND} & Shifted IND                  & Original EXD                  & Shifted EXD                 & \parbox{1.5cm}{ \centering    synthetical                           \\ tests} & \parbox{1.5cm}{ \centering Average}    \\ \midrule
            FedAvg            & 73.52 $_{\pm 0.22}$                               & 21.82 $_{\pm 0.09}$         & 73.06 $_{\pm 0.35}$          & 21.60 $_{\pm 0.15}$          & 40.54 $_{\pm 0.05}$                & 47.51           \\      
            FedAvg+FT         & 66.11 $_{\pm 0.29}$                               & 21.69 $_{\pm 0.04}$         & 48.58 $_{\pm 0.48}$          & 14.23 $_{\pm 0.06}$          & 37.68 $_{\pm 0.09}$                & 37.66       \\
			FedRoD            & 65.95 $_{\pm 0.76}$                               & 21.04 $_{\pm 0.45}$         & 46.53 $_{\pm 0.31}$          & 13.63 $_{\pm 0.22}$          & 36.58 $_{\pm 0.42}$                & 36.75        \\
			FedRep            & 65.36 $_{\pm 0.46}$                               & 21.10 $_{\pm 0.10}$         & 44.22 $_{\pm 0.37}$          & 13.27 $_{\pm 0.19}$          & 35.99 $_{\pm 0.06}$               & 35.99       \\
            kNN-Per           & 76.81 $_{\pm 0.01}$                               & 24.57 $_{\pm 0.08}$         & 69.16 $_{\pm 0.11}$          & 21.00 $_{\pm 0.02}$          & 47.90 $_{\pm 0.01}$                & 47.89        \\
            MEMO              & 64.37 $_{\pm 0.43}$                               & 18.36 $_{\pm 0.23}$         & 48.00 $_{\pm 0.31}$          & 11.69 $_{\pm 0.12}$           & 34.27 $_{\pm 0.14}$                & 35.34       \\
			TTT               & 64.33 $_{\pm 0.50}$                               & 24.04 $_{\pm 0.31}$         & 47.66 $_{\pm 0.29}$          & 15.67 $_{\pm 0.18}$           & 36.34 $_{\pm 0.017}$                & 37.61      \\
            FedTHE            & 81.74 $_{\pm 0.06}$                               & 25.28 $_{\pm 0.00}$         & 69.91 $_{\pm 0.65}$          & 20.77 $_{\pm 0.32}$          & 48.97 $_{\pm 0.10}$                & 49.33      \\      
            \midrule
			BTFL(Ours)        & 78.58 $_{\pm 0.11}$                               & 24.00 $_{\pm 0.13}$          & 73.92 $_{\pm 0.17}$         & 22.19 $_{\pm 0.16}$          & 49.13 $_{\pm 0.16}$               & \textbf{49.55}         \\
			\bottomrule
		\end{tabular}%
	}
\end{table}

\subsection{Visualization of head interpolation coefficient.}
Here, we analyze and visualize the distribution of interpolation coefficients $e$ for BTFL, with Figure \ref{fig:ensemble} showing the proportion distribution of $e$ (coefficient for global head) on range $[0,1]$ across different test distributions. Note that the last subplot shows the averaged result of the first four subplots to compare with the subplot 'Synthetical'. We can see that:
\begin{itemize}
    \item For original/shifted IND distribution. The coefficient clearly tends to utilize the local head for original set: the proportion of the local head exceeds that of the local head by more than 95\%, hence the personalized knowledge will take a dominant position. However, for shifted samples, $e$ tends to be less concentrated due to feature divergence.
    
    \item for original/shifted EXD distribution. Compared to the IND scenario, the proportion of the global head has increased. Also, we find the result of original and co-variate shifted samples share similar pattern, this might be owing to global DLE is more stable and less affected by feature shifts. 
    
    \item For Synthetical distribution. The coefficient distribution of the Synthetical dataset (left) forms a more concentrated Bimodal shape than the average distribution of the above four (right), indicating that our method seeks for a stable interpolation strategy to deal with heterogeneous data flow.
\end{itemize}


\begin{thebibliography}{53}


\ifx \showCODEN    \undefined \def \showCODEN     #1{\unskip}     \fi
\ifx \showDOI      \undefined \def \showDOI       #1{#1}\fi
\ifx \showISBNx    \undefined \def \showISBNx     #1{\unskip}     \fi
\ifx \showISBNxiii \undefined \def \showISBNxiii  #1{\unskip}     \fi
\ifx \showISSN     \undefined \def \showISSN      #1{\unskip}     \fi
\ifx \showLCCN     \undefined \def \showLCCN      #1{\unskip}     \fi
\ifx \shownote     \undefined \def \shownote      #1{#1}          \fi
\ifx \showarticletitle \undefined \def \showarticletitle #1{#1}   \fi
\ifx \showURL      \undefined \def \showURL       {\relax}        \fi
\providecommand\bibfield[2]{#2}
\providecommand\bibinfo[2]{#2}
\providecommand\natexlab[1]{#1}
\providecommand\showeprint[2][]{arXiv:#2}

\bibitem[Abdullah et~al\mbox{.}(2024)]%
        {abdullah2024leveraging}
\bibfield{author}{\bibinfo{person}{Abdullah~A Abdullah}, \bibinfo{person}{Masoud~M Hassan}, {and} \bibinfo{person}{Yaseen~T Mustafa}.} \bibinfo{year}{2024}\natexlab{}.
\newblock \showarticletitle{Leveraging Bayesian deep learning and ensemble methods for uncertainty quantification in image classification: A ranking-based approach}.
\newblock \bibinfo{journal}{\emph{Heliyon}} \bibinfo{volume}{10}, \bibinfo{number}{2} (\bibinfo{year}{2024}).
\newblock


\bibitem[Alexandari et~al\mbox{.}(2020)]%
        {alexandari2020maximum}
\bibfield{author}{\bibinfo{person}{Amr Alexandari}, \bibinfo{person}{Anshul Kundaje}, {and} \bibinfo{person}{Avanti Shrikumar}.} \bibinfo{year}{2020}\natexlab{}.
\newblock \showarticletitle{Maximum likelihood with bias-corrected calibration is hard-to-beat at label shift adaptation}. In \bibinfo{booktitle}{\emph{International Conference on Machine Learning}}. PMLR, \bibinfo{pages}{222--232}.
\newblock


\bibitem[Azizzadenesheli et~al\mbox{.}(2019)]%
        {azizzadenesheli2019regularized}
\bibfield{author}{\bibinfo{person}{Kamyar Azizzadenesheli}, \bibinfo{person}{Anqi Liu}, \bibinfo{person}{Fanny Yang}, {and} \bibinfo{person}{Animashree Anandkumar}.} \bibinfo{year}{2019}\natexlab{}.
\newblock \showarticletitle{Regularized learning for domain adaptation under label shifts}.
\newblock \bibinfo{journal}{\emph{arXiv preprint arXiv:1903.09734}} (\bibinfo{year}{2019}).
\newblock


\bibitem[Bao et~al\mbox{.}(2024)]%
        {bao2024adaptive}
\bibfield{author}{\bibinfo{person}{Wenxuan Bao}, \bibinfo{person}{Tianxin Wei}, \bibinfo{person}{Haohan Wang}, {and} \bibinfo{person}{Jingrui He}.} \bibinfo{year}{2024}\natexlab{}.
\newblock \showarticletitle{Adaptive Test-Time Personalization for Federated Learning}.
\newblock \bibinfo{journal}{\emph{Advances in Neural Information Processing Systems}}  \bibinfo{volume}{36} (\bibinfo{year}{2024}).
\newblock


\bibitem[Caldarola et~al\mbox{.}(2022)]%
        {caldarola2022improving}
\bibfield{author}{\bibinfo{person}{Debora Caldarola}, \bibinfo{person}{Barbara Caputo}, {and} \bibinfo{person}{Marco Ciccone}.} \bibinfo{year}{2022}\natexlab{}.
\newblock \bibinfo{title}{Improving Generalization in Federated Learning by Seeking Flat Minima}.
\newblock
\newblock
\showeprint[arxiv]{2203.11834}~[cs.LG]


\bibitem[Chen and Chao(2020)]%
        {chen2020fedbe}
\bibfield{author}{\bibinfo{person}{Hong-You Chen} {and} \bibinfo{person}{Wei-Lun Chao}.} \bibinfo{year}{2020}\natexlab{}.
\newblock \showarticletitle{Fedbe: Making bayesian model ensemble applicable to federated learning}.
\newblock \bibinfo{journal}{\emph{arXiv preprint arXiv:2009.01974}} (\bibinfo{year}{2020}).
\newblock


\bibitem[Chen and Chao(2022)]%
        {chen2022on}
\bibfield{author}{\bibinfo{person}{Hong-You Chen} {and} \bibinfo{person}{Wei-Lun Chao}.} \bibinfo{year}{2022}\natexlab{}.
\newblock \showarticletitle{On Bridging Generic and Personalized Federated Learning for Image Classification}. In \bibinfo{booktitle}{\emph{International Conference on Learning Representations}}.
\newblock
\urldef\tempurl%
\url{https://openreview.net/forum?id=I1hQbx10Kxn}
\showURL{%
\tempurl}


\bibitem[Chrabaszcz et~al\mbox{.}(2017)]%
        {chrabaszcz2017downsampled}
\bibfield{author}{\bibinfo{person}{Patryk Chrabaszcz}, \bibinfo{person}{Ilya Loshchilov}, {and} \bibinfo{person}{Frank Hutter}.} \bibinfo{year}{2017}\natexlab{}.
\newblock \showarticletitle{A downsampled variant of imagenet as an alternative to the cifar datasets}.
\newblock \bibinfo{journal}{\emph{arXiv preprint arXiv:1707.08819}} (\bibinfo{year}{2017}).
\newblock


\bibitem[Deng et~al\mbox{.}(2009)]%
        {deng2009imagenet}
\bibfield{author}{\bibinfo{person}{Jia Deng}, \bibinfo{person}{Wei Dong}, \bibinfo{person}{Richard Socher}, \bibinfo{person}{Li-Jia Li}, \bibinfo{person}{Kai Li}, {and} \bibinfo{person}{Li Fei-Fei}.} \bibinfo{year}{2009}\natexlab{}.
\newblock \showarticletitle{Imagenet: A large-scale hierarchical image database}. In \bibinfo{booktitle}{\emph{2009 IEEE conference on computer vision and pattern recognition}}. Ieee, \bibinfo{pages}{248--255}.
\newblock


\bibitem[Dubs(1942)]%
        {dubs1942principle}
\bibfield{author}{\bibinfo{person}{Homer~H Dubs}.} \bibinfo{year}{1942}\natexlab{}.
\newblock \showarticletitle{The principle of insufficient reason}.
\newblock \bibinfo{journal}{\emph{Philosophy of Science}} \bibinfo{volume}{9}, \bibinfo{number}{2} (\bibinfo{year}{1942}), \bibinfo{pages}{123--131}.
\newblock


\bibitem[Fischer et~al\mbox{.}(2024)]%
        {fischer2024federated}
\bibfield{author}{\bibinfo{person}{John Fischer}, \bibinfo{person}{Marko Orescanin}, \bibinfo{person}{Justin Loomis}, {and} \bibinfo{person}{Patrick McClure}.} \bibinfo{year}{2024}\natexlab{}.
\newblock \showarticletitle{Federated Bayesian Deep Learning: The Application of Statistical Aggregation Methods to Bayesian Models}.
\newblock \bibinfo{journal}{\emph{arXiv preprint arXiv:2403.15263}} (\bibinfo{year}{2024}).
\newblock


\bibitem[Hassani et~al\mbox{.}(2021)]%
        {hassani2021escaping}
\bibfield{author}{\bibinfo{person}{Ali Hassani}, \bibinfo{person}{Steven Walton}, \bibinfo{person}{Nikhil Shah}, \bibinfo{person}{Abulikemu Abuduweili}, \bibinfo{person}{Jiachen Li}, {and} \bibinfo{person}{Humphrey Shi}.} \bibinfo{year}{2021}\natexlab{}.
\newblock \showarticletitle{Escaping the big data paradigm with compact transformers}.
\newblock \bibinfo{journal}{\emph{arXiv preprint arXiv:2104.05704}} (\bibinfo{year}{2021}).
\newblock


\bibitem[He et~al\mbox{.}(2016)]%
        {he2016deep}
\bibfield{author}{\bibinfo{person}{Kaiming He}, \bibinfo{person}{Xiangyu Zhang}, \bibinfo{person}{Shaoqing Ren}, {and} \bibinfo{person}{Jian Sun}.} \bibinfo{year}{2016}\natexlab{}.
\newblock \showarticletitle{Deep residual learning for image recognition}. In \bibinfo{booktitle}{\emph{Proceedings of the IEEE conference on computer vision and pattern recognition}}. \bibinfo{pages}{770--778}.
\newblock


\bibitem[Hendrycks et~al\mbox{.}(2021)]%
        {hendrycks2021many}
\bibfield{author}{\bibinfo{person}{Dan Hendrycks}, \bibinfo{person}{Steven Basart}, \bibinfo{person}{Norman Mu}, \bibinfo{person}{Saurav Kadavath}, \bibinfo{person}{Frank Wang}, \bibinfo{person}{Evan Dorundo}, \bibinfo{person}{Rahul Desai}, \bibinfo{person}{Tyler Zhu}, \bibinfo{person}{Samyak Parajuli}, \bibinfo{person}{Mike Guo}, {et~al\mbox{.}}} \bibinfo{year}{2021}\natexlab{}.
\newblock \showarticletitle{The many faces of robustness: A critical analysis of out-of-distribution generalization}. In \bibinfo{booktitle}{\emph{Proceedings of the IEEE/CVF International Conference on Computer Vision}}. \bibinfo{pages}{8340--8349}.
\newblock


\bibitem[Hendrycks and Dietterich(2018)]%
        {hendrycks2018benchmarking}
\bibfield{author}{\bibinfo{person}{Dan Hendrycks} {and} \bibinfo{person}{Thomas~G Dietterich}.} \bibinfo{year}{2018}\natexlab{}.
\newblock \showarticletitle{Benchmarking neural network robustness to common corruptions and surface variations}.
\newblock \bibinfo{journal}{\emph{arXiv preprint arXiv:1807.01697}} (\bibinfo{year}{2018}).
\newblock


\bibitem[Hsieh et~al\mbox{.}(2020)]%
        {hsieh2020non}
\bibfield{author}{\bibinfo{person}{Kevin Hsieh}, \bibinfo{person}{Amar Phanishayee}, \bibinfo{person}{Onur Mutlu}, {and} \bibinfo{person}{Phillip Gibbons}.} \bibinfo{year}{2020}\natexlab{}.
\newblock \showarticletitle{The non-iid data quagmire of decentralized machine learning}. In \bibinfo{booktitle}{\emph{International Conference on Machine Learning}}. PMLR, \bibinfo{pages}{4387--4398}.
\newblock


\bibitem[Hsu et~al\mbox{.}(2019)]%
        {hsu2019measuring}
\bibfield{author}{\bibinfo{person}{Tzu-Ming~Harry Hsu}, \bibinfo{person}{Hang Qi}, {and} \bibinfo{person}{Matthew Brown}.} \bibinfo{year}{2019}\natexlab{}.
\newblock \showarticletitle{Measuring the effects of non-identical data distribution for federated visual classification}.
\newblock \bibinfo{journal}{\emph{arXiv preprint arXiv:1909.06335}} (\bibinfo{year}{2019}).
\newblock


\bibitem[Iwasawa and Matsuo(2021)]%
        {iwasawa2021test}
\bibfield{author}{\bibinfo{person}{Yusuke Iwasawa} {and} \bibinfo{person}{Yutaka Matsuo}.} \bibinfo{year}{2021}\natexlab{}.
\newblock \showarticletitle{Test-time classifier adjustment module for model-agnostic domain generalization}.
\newblock \bibinfo{journal}{\emph{Advances in Neural Information Processing Systems}}  \bibinfo{volume}{34} (\bibinfo{year}{2021}).
\newblock


\bibitem[Jiang and Lin(2022)]%
        {jiang2022test}
\bibfield{author}{\bibinfo{person}{Liangze Jiang} {and} \bibinfo{person}{Tao Lin}.} \bibinfo{year}{2022}\natexlab{}.
\newblock \showarticletitle{Test-time robust personalization for federated learning}.
\newblock \bibinfo{journal}{\emph{arXiv preprint arXiv:2205.10920}} (\bibinfo{year}{2022}).
\newblock


\bibitem[Kingma and Ba(2014)]%
        {kingma2014adam}
\bibfield{author}{\bibinfo{person}{Diederik~P Kingma} {and} \bibinfo{person}{Jimmy Ba}.} \bibinfo{year}{2014}\natexlab{}.
\newblock \showarticletitle{Adam: A method for stochastic optimization}.
\newblock \bibinfo{journal}{\emph{arXiv preprint arXiv:1412.6980}} (\bibinfo{year}{2014}).
\newblock


\bibitem[Krizhevsky et~al\mbox{.}(2009)]%
        {krizhevsky2009learning}
\bibfield{author}{\bibinfo{person}{Alex Krizhevsky}, \bibinfo{person}{Geoffrey Hinton}, {et~al\mbox{.}}} \bibinfo{year}{2009}\natexlab{}.
\newblock \showarticletitle{Learning multiple layers of features from tiny images}.
\newblock  (\bibinfo{year}{2009}).
\newblock


\bibitem[LeCun et~al\mbox{.}(1998)]%
        {lecun1998gradient}
\bibfield{author}{\bibinfo{person}{Yann LeCun}, \bibinfo{person}{L{\'e}on Bottou}, \bibinfo{person}{Yoshua Bengio}, {and} \bibinfo{person}{Patrick Haffner}.} \bibinfo{year}{1998}\natexlab{}.
\newblock \showarticletitle{Gradient-based learning applied to document recognition}.
\newblock \bibinfo{journal}{\emph{Proc. IEEE}} \bibinfo{volume}{86}, \bibinfo{number}{11} (\bibinfo{year}{1998}), \bibinfo{pages}{2278--2324}.
\newblock


\bibitem[Lee et~al\mbox{.}(2024)]%
        {lee2024entropy}
\bibfield{author}{\bibinfo{person}{Jonghyun Lee}, \bibinfo{person}{Dahuin Jung}, \bibinfo{person}{Saehyung Lee}, \bibinfo{person}{Junsung Park}, \bibinfo{person}{Juhyeon Shin}, \bibinfo{person}{Uiwon Hwang}, {and} \bibinfo{person}{Sungroh Yoon}.} \bibinfo{year}{2024}\natexlab{}.
\newblock \showarticletitle{Entropy is not enough for test-time adaptation: From the perspective of disentangled factors}.
\newblock \bibinfo{journal}{\emph{arXiv preprint arXiv:2403.07366}} (\bibinfo{year}{2024}).
\newblock


\bibitem[Liu et~al\mbox{.}(2023)]%
        {liu2023beyond}
\bibfield{author}{\bibinfo{person}{Bingyan Liu}, \bibinfo{person}{Yifeng Cai}, \bibinfo{person}{Hongzhe Bi}, \bibinfo{person}{Ziqi Zhang}, \bibinfo{person}{Ding Li}, \bibinfo{person}{Yao Guo}, {and} \bibinfo{person}{Xiangqun Chen}.} \bibinfo{year}{2023}\natexlab{}.
\newblock \showarticletitle{Beyond Fine-Tuning: Efficient and Effective Fed-Tuning for Mobile/Web Users}. In \bibinfo{booktitle}{\emph{Proceedings of the ACM Web Conference 2023}}. \bibinfo{pages}{2863--2873}.
\newblock


\bibitem[Liu et~al\mbox{.}(2021a)]%
        {liu2021distfl}
\bibfield{author}{\bibinfo{person}{Bingyan Liu}, \bibinfo{person}{Yifeng Cai}, \bibinfo{person}{Ziqi Zhang}, \bibinfo{person}{Yuanchun Li}, \bibinfo{person}{Leye Wang}, \bibinfo{person}{Ding Li}, \bibinfo{person}{Yao Guo}, {and} \bibinfo{person}{Xiangqun Chen}.} \bibinfo{year}{2021}\natexlab{a}.
\newblock \showarticletitle{DistFL: Distribution-aware Federated Learning for Mobile Scenarios}.
\newblock \bibinfo{journal}{\emph{Proceedings of the ACM on Interactive, Mobile, Wearable and Ubiquitous Technologies}} \bibinfo{volume}{5}, \bibinfo{number}{4} (\bibinfo{year}{2021}), \bibinfo{pages}{1--26}.
\newblock


\bibitem[Liu et~al\mbox{.}(2021b)]%
        {liu2021pfa}
\bibfield{author}{\bibinfo{person}{Bingyan Liu}, \bibinfo{person}{Yao Guo}, {and} \bibinfo{person}{Xiangqun Chen}.} \bibinfo{year}{2021}\natexlab{b}.
\newblock \showarticletitle{PFA: Privacy-preserving Federated Adaptation for Effective Model Personalization}. In \bibinfo{booktitle}{\emph{Proceedings of the Web Conference 2021}}. \bibinfo{pages}{923--934}.
\newblock


\bibitem[Liu et~al\mbox{.}(2024)]%
        {liu2024recent}
\bibfield{author}{\bibinfo{person}{Bingyan Liu}, \bibinfo{person}{Nuoyan Lv}, \bibinfo{person}{Yuanchun Guo}, {and} \bibinfo{person}{Yawen Li}.} \bibinfo{year}{2024}\natexlab{}.
\newblock \showarticletitle{Recent advances on federated learning: A systematic survey}.
\newblock \bibinfo{journal}{\emph{Neurocomputing}} (\bibinfo{year}{2024}), \bibinfo{pages}{128019}.
\newblock


\bibitem[McMahan et~al\mbox{.}(2017)]%
        {mcmahan2017communication}
\bibfield{author}{\bibinfo{person}{Brendan McMahan}, \bibinfo{person}{Eider Moore}, \bibinfo{person}{Daniel Ramage}, \bibinfo{person}{Seth Hampson}, {and} \bibinfo{person}{Blaise~Aguera y Arcas}.} \bibinfo{year}{2017}\natexlab{}.
\newblock \showarticletitle{Communication-efficient learning of deep networks from decentralized data}. In \bibinfo{booktitle}{\emph{Artificial Intelligence and Statistics}}. PMLR, \bibinfo{pages}{1273--1282}.
\newblock


\bibitem[Mentzer et~al\mbox{.}(2023)]%
        {mentzer2023finite}
\bibfield{author}{\bibinfo{person}{Fabian Mentzer}, \bibinfo{person}{David Minnen}, \bibinfo{person}{Eirikur Agustsson}, {and} \bibinfo{person}{Michael Tschannen}.} \bibinfo{year}{2023}\natexlab{}.
\newblock \showarticletitle{Finite scalar quantization: Vq-vae made simple}.
\newblock \bibinfo{journal}{\emph{arXiv preprint arXiv:2309.15505}} (\bibinfo{year}{2023}).
\newblock


\bibitem[Nguyen et~al\mbox{.}(2022)]%
        {nguyen2022fedsr}
\bibfield{author}{\bibinfo{person}{A~Tuan Nguyen}, \bibinfo{person}{Philip Torr}, {and} \bibinfo{person}{Ser~Nam Lim}.} \bibinfo{year}{2022}\natexlab{}.
\newblock \showarticletitle{Fedsr: A simple and effective domain generalization method for federated learning}.
\newblock \bibinfo{journal}{\emph{Advances in Neural Information Processing Systems}}  \bibinfo{volume}{35} (\bibinfo{year}{2022}), \bibinfo{pages}{38831--38843}.
\newblock


\bibitem[Qu et~al\mbox{.}(2022)]%
        {qu2022generalized}
\bibfield{author}{\bibinfo{person}{Zhe Qu}, \bibinfo{person}{Xingyu Li}, \bibinfo{person}{Rui Duan}, \bibinfo{person}{Yao Liu}, \bibinfo{person}{Bo Tang}, {and} \bibinfo{person}{Zhuo Lu}.} \bibinfo{year}{2022}\natexlab{}.
\newblock \showarticletitle{Generalized federated learning via sharpness aware minimization}. In \bibinfo{booktitle}{\emph{International conference on machine learning}}. PMLR, \bibinfo{pages}{18250--18280}.
\newblock


\bibitem[Recht et~al\mbox{.}(2018)]%
        {recht2018cifar}
\bibfield{author}{\bibinfo{person}{Benjamin Recht}, \bibinfo{person}{Rebecca Roelofs}, \bibinfo{person}{Ludwig Schmidt}, {and} \bibinfo{person}{Vaishaal Shankar}.} \bibinfo{year}{2018}\natexlab{}.
\newblock \showarticletitle{Do CIFAR-10 classifiers generalize to CIFAR-10?}
\newblock \bibinfo{journal}{\emph{arXiv preprint arXiv:1806.00451}} (\bibinfo{year}{2018}).
\newblock


\bibitem[Reisizadeh et~al\mbox{.}(2020)]%
        {reisizadeh2020robust}
\bibfield{author}{\bibinfo{person}{Amirhossein Reisizadeh}, \bibinfo{person}{Farzan Farnia}, \bibinfo{person}{Ramtin Pedarsani}, {and} \bibinfo{person}{Ali Jadbabaie}.} \bibinfo{year}{2020}\natexlab{}.
\newblock \showarticletitle{Robust federated learning: The case of affine distribution shifts}.
\newblock \bibinfo{journal}{\emph{Advances in Neural Information Processing Systems}}  \bibinfo{volume}{33} (\bibinfo{year}{2020}), \bibinfo{pages}{21554--21565}.
\newblock


\bibitem[Saerens et~al\mbox{.}(2002)]%
        {saerens2002adjusting}
\bibfield{author}{\bibinfo{person}{Marco Saerens}, \bibinfo{person}{Patrice Latinne}, {and} \bibinfo{person}{Christine Decaestecker}.} \bibinfo{year}{2002}\natexlab{}.
\newblock \showarticletitle{Adjusting the outputs of a classifier to new a priori probabilities: a simple procedure}.
\newblock \bibinfo{journal}{\emph{Neural computation}} \bibinfo{volume}{14}, \bibinfo{number}{1} (\bibinfo{year}{2002}), \bibinfo{pages}{21--41}.
\newblock


\bibitem[Seligmann et~al\mbox{.}(2024)]%
        {seligmann2024beyond}
\bibfield{author}{\bibinfo{person}{Florian Seligmann}, \bibinfo{person}{Philipp Becker}, \bibinfo{person}{Michael Volpp}, {and} \bibinfo{person}{Gerhard Neumann}.} \bibinfo{year}{2024}\natexlab{}.
\newblock \showarticletitle{Beyond Deep Ensembles: A Large-Scale Evaluation of Bayesian Deep Learning under Distribution Shift}.
\newblock \bibinfo{journal}{\emph{Advances in Neural Information Processing Systems}}  \bibinfo{volume}{36} (\bibinfo{year}{2024}).
\newblock


\bibitem[Shi et~al\mbox{.}(2021)]%
        {shi2021fed}
\bibfield{author}{\bibinfo{person}{Naichen Shi}, \bibinfo{person}{Fan Lai}, \bibinfo{person}{Raed~Al Kontar}, {and} \bibinfo{person}{Mosharaf Chowdhury}.} \bibinfo{year}{2021}\natexlab{}.
\newblock \showarticletitle{Fed-ensemble: Improving generalization through model ensembling in federated learning}.
\newblock \bibinfo{journal}{\emph{arXiv preprint arXiv:2107.10663}} (\bibinfo{year}{2021}).
\newblock


\bibitem[Staton et~al\mbox{.}(2018)]%
        {staton2018beta}
\bibfield{author}{\bibinfo{person}{Sam Staton}, \bibinfo{person}{Dario Stein}, \bibinfo{person}{Hongseok Yang}, \bibinfo{person}{Nathanael~L Ackerman}, \bibinfo{person}{Cameron~E Freer}, {and} \bibinfo{person}{Daniel~M Roy}.} \bibinfo{year}{2018}\natexlab{}.
\newblock \showarticletitle{The Beta-Bernoulli process and algebraic effects}.
\newblock \bibinfo{journal}{\emph{arXiv preprint arXiv:1802.09598}} (\bibinfo{year}{2018}).
\newblock


\bibitem[Sun et~al\mbox{.}(2020)]%
        {sun2020test}
\bibfield{author}{\bibinfo{person}{Yu Sun}, \bibinfo{person}{Xiaolong Wang}, \bibinfo{person}{Zhuang Liu}, \bibinfo{person}{John Miller}, \bibinfo{person}{Alexei Efros}, {and} \bibinfo{person}{Moritz Hardt}.} \bibinfo{year}{2020}\natexlab{}.
\newblock \showarticletitle{Test-time training with self-supervision for generalization under distribution shifts}. In \bibinfo{booktitle}{\emph{International Conference on Machine Learning}}. PMLR, \bibinfo{pages}{9229--9248}.
\newblock


\bibitem[Tan et~al\mbox{.}(2023)]%
        {tan2023taming}
\bibfield{author}{\bibinfo{person}{Yue Tan}, \bibinfo{person}{Chen Chen}, \bibinfo{person}{Weiming Zhuang}, \bibinfo{person}{Xin Dong}, \bibinfo{person}{Lingjuan Lyu}, {and} \bibinfo{person}{Guodong Long}.} \bibinfo{year}{2023}\natexlab{}.
\newblock \showarticletitle{Taming heterogeneity to deal with test-time shift in federated learning}. In \bibinfo{booktitle}{\emph{International Workshop on Federated Learning for Distributed Data Mining}}.
\newblock


\bibitem[Venkateswara et~al\mbox{.}(2017)]%
        {venkateswara2017deep}
\bibfield{author}{\bibinfo{person}{Hemanth Venkateswara}, \bibinfo{person}{Jose Eusebio}, \bibinfo{person}{Shayok Chakraborty}, {and} \bibinfo{person}{Sethuraman Panchanathan}.} \bibinfo{year}{2017}\natexlab{}.
\newblock \showarticletitle{Deep hashing network for unsupervised domain adaptation}. In \bibinfo{booktitle}{\emph{Proceedings of the IEEE conference on computer vision and pattern recognition}}. \bibinfo{pages}{5018--5027}.
\newblock


\bibitem[Wang et~al\mbox{.}(2020)]%
        {wang2020tent}
\bibfield{author}{\bibinfo{person}{Dequan Wang}, \bibinfo{person}{Evan Shelhamer}, \bibinfo{person}{Shaoteng Liu}, \bibinfo{person}{Bruno Olshausen}, {and} \bibinfo{person}{Trevor Darrell}.} \bibinfo{year}{2020}\natexlab{}.
\newblock \showarticletitle{Tent: Fully test-time adaptation by entropy minimization}.
\newblock \bibinfo{journal}{\emph{arXiv preprint arXiv:2006.10726}} (\bibinfo{year}{2020}).
\newblock


\bibitem[Wang et~al\mbox{.}(2024a)]%
        {wang2024feddse}
\bibfield{author}{\bibinfo{person}{Haozhao Wang}, \bibinfo{person}{Yabo Jia}, \bibinfo{person}{Meng Zhang}, \bibinfo{person}{Qinghao Hu}, \bibinfo{person}{Hao Ren}, \bibinfo{person}{Peng Sun}, \bibinfo{person}{Yonggang Wen}, {and} \bibinfo{person}{Tianwei Zhang}.} \bibinfo{year}{2024}\natexlab{a}.
\newblock \showarticletitle{FedDSE: Distribution-aware Sub-model Extraction for Federated Learning over Resource-constrained Devices}. In \bibinfo{booktitle}{\emph{Proceedings of the ACM on Web Conference 2024}}. \bibinfo{pages}{2902--2913}.
\newblock


\bibitem[Wang et~al\mbox{.}(2023)]%
        {wang2023dafkd}
\bibfield{author}{\bibinfo{person}{Haozhao Wang}, \bibinfo{person}{Yichen Li}, \bibinfo{person}{Wenchao Xu}, \bibinfo{person}{Ruixuan Li}, \bibinfo{person}{Yufeng Zhan}, {and} \bibinfo{person}{Zhigang Zeng}.} \bibinfo{year}{2023}\natexlab{}.
\newblock \showarticletitle{Dafkd: Domain-aware federated knowledge distillation}. In \bibinfo{booktitle}{\emph{Proceedings of the IEEE/CVF conference on computer vision and pattern recognition}}. \bibinfo{pages}{20412--20421}.
\newblock


\bibitem[Wang et~al\mbox{.}(2024c)]%
        {wang2024fedcda}
\bibfield{author}{\bibinfo{person}{Haozhao Wang}, \bibinfo{person}{Haoran Xu}, \bibinfo{person}{Yichen Li}, \bibinfo{person}{Yuan Xu}, \bibinfo{person}{Ruixuan Li}, {and} \bibinfo{person}{Tianwei Zhang}.} \bibinfo{year}{2024}\natexlab{c}.
\newblock \showarticletitle{FedCDA: Federated Learning with Cross-rounds Divergence-aware Aggregation}. In \bibinfo{booktitle}{\emph{The Twelfth International Conference on Learning Representations}}.
\newblock


\bibitem[Wang et~al\mbox{.}(2024d)]%
        {wang2024fednlr}
\bibfield{author}{\bibinfo{person}{Haozhao Wang}, \bibinfo{person}{Peirong Zheng}, \bibinfo{person}{Xingshuo Han}, \bibinfo{person}{Wenchao Xu}, \bibinfo{person}{Ruixuan Li}, {and} \bibinfo{person}{Tianwei Zhang}.} \bibinfo{year}{2024}\natexlab{d}.
\newblock \showarticletitle{Fednlr: Federated learning with neuron-wise learning rates}. In \bibinfo{booktitle}{\emph{Proceedings of the 30th ACM SIGKDD Conference on Knowledge Discovery and Data Mining}}. \bibinfo{pages}{3069--3080}.
\newblock


\bibitem[Wang et~al\mbox{.}(2019)]%
        {wang2019federated}
\bibfield{author}{\bibinfo{person}{Kangkang Wang}, \bibinfo{person}{Rajiv Mathews}, \bibinfo{person}{Chlo{\'e} Kiddon}, \bibinfo{person}{Hubert Eichner}, \bibinfo{person}{Fran{\c{c}}oise Beaufays}, {and} \bibinfo{person}{Daniel Ramage}.} \bibinfo{year}{2019}\natexlab{}.
\newblock \showarticletitle{Federated evaluation of on-device personalization}.
\newblock \bibinfo{journal}{\emph{arXiv preprint arXiv:1910.10252}} (\bibinfo{year}{2019}).
\newblock


\bibitem[Wang et~al\mbox{.}(2024b)]%
        {wang2024traceable}
\bibfield{author}{\bibinfo{person}{Qiang Wang}, \bibinfo{person}{Bingyan Liu}, {and} \bibinfo{person}{Yawen Li}.} \bibinfo{year}{2024}\natexlab{b}.
\newblock \showarticletitle{Traceable Federated Continual Learning}. In \bibinfo{booktitle}{\emph{Proceedings of the IEEE/CVF Conference on Computer Vision and Pattern Recognition}}. \bibinfo{pages}{12872--12881}.
\newblock


\bibitem[Wilson and Izmailov(2020)]%
        {wilson2020bayesian}
\bibfield{author}{\bibinfo{person}{Andrew~G Wilson} {and} \bibinfo{person}{Pavel Izmailov}.} \bibinfo{year}{2020}\natexlab{}.
\newblock \showarticletitle{Bayesian deep learning and a probabilistic perspective of generalization}.
\newblock \bibinfo{journal}{\emph{Advances in neural information processing systems}}  \bibinfo{volume}{33} (\bibinfo{year}{2020}), \bibinfo{pages}{4697--4708}.
\newblock


\bibitem[Wortsman et~al\mbox{.}(2022)]%
        {wortsman2022robust}
\bibfield{author}{\bibinfo{person}{Mitchell Wortsman}, \bibinfo{person}{Gabriel Ilharco}, \bibinfo{person}{Jong~Wook Kim}, \bibinfo{person}{Mike Li}, \bibinfo{person}{Simon Kornblith}, \bibinfo{person}{Rebecca Roelofs}, \bibinfo{person}{Raphael~Gontijo Lopes}, \bibinfo{person}{Hannaneh Hajishirzi}, \bibinfo{person}{Ali Farhadi}, \bibinfo{person}{Hongseok Namkoong}, {et~al\mbox{.}}} \bibinfo{year}{2022}\natexlab{}.
\newblock \showarticletitle{Robust fine-tuning of zero-shot models}. In \bibinfo{booktitle}{\emph{Proceedings of the IEEE/CVF conference on computer vision and pattern recognition}}. \bibinfo{pages}{7959--7971}.
\newblock


\bibitem[Wu and He(2018)]%
        {wu2018group}
\bibfield{author}{\bibinfo{person}{Yuxin Wu} {and} \bibinfo{person}{Kaiming He}.} \bibinfo{year}{2018}\natexlab{}.
\newblock \showarticletitle{Group normalization}. In \bibinfo{booktitle}{\emph{Proceedings of the European conference on computer vision (ECCV)}}. \bibinfo{pages}{3--19}.
\newblock


\bibitem[Yurochkin et~al\mbox{.}(2019)]%
        {yurochkin2019bayesian}
\bibfield{author}{\bibinfo{person}{Mikhail Yurochkin}, \bibinfo{person}{Mayank Agarwal}, \bibinfo{person}{Soumya Ghosh}, \bibinfo{person}{Kristjan Greenewald}, \bibinfo{person}{Nghia Hoang}, {and} \bibinfo{person}{Yasaman Khazaeni}.} \bibinfo{year}{2019}\natexlab{}.
\newblock \showarticletitle{Bayesian nonparametric federated learning of neural networks}. In \bibinfo{booktitle}{\emph{International conference on machine learning}}. PMLR, \bibinfo{pages}{7252--7261}.
\newblock


\bibitem[Zeng and Van~den Broeck(2024)]%
        {zeng2024collapsed}
\bibfield{author}{\bibinfo{person}{Zhe Zeng} {and} \bibinfo{person}{Guy Van~den Broeck}.} \bibinfo{year}{2024}\natexlab{}.
\newblock \showarticletitle{Collapsed inference for bayesian deep learning}.
\newblock \bibinfo{journal}{\emph{Advances in Neural Information Processing Systems}}  \bibinfo{volume}{36} (\bibinfo{year}{2024}).
\newblock


\bibitem[Zhang et~al\mbox{.}(2021)]%
        {zhang2021memo}
\bibfield{author}{\bibinfo{person}{Marvin Zhang}, \bibinfo{person}{Sergey Levine}, {and} \bibinfo{person}{Chelsea Finn}.} \bibinfo{year}{2021}\natexlab{}.
\newblock \showarticletitle{MEMO: Test Time Robustness via Adaptation and Augmentation}.
\newblock \bibinfo{journal}{\emph{arXiv preprint arXiv:2110.09506}} (\bibinfo{year}{2021}).
\newblock


\end{thebibliography}
\end{document}